%% file: main.tex
\documentclass[10pt,journal,compsoc]{IEEEtran}
\usepackage{graphicx}
\usepackage{subfig}
\usepackage{multirow}
\usepackage{amssymb}
\usepackage[fleqn]{amsmath}
\usepackage{booktabs}
\usepackage{tabularx}
\usepackage{xcolor}
\usepackage{color}

\newcommand{\eg}{\textit{e.g. }}
\newcommand{\ie}{\textit{i.e. }}
\newcommand{\YL}[1]{{\color{black}#1}}
\newcommand{\gl}[1]{{\color{black}#1}}
\newcommand{\cl}[1]{{\color{black}#1}}
\newcolumntype{L}[1]{>{\raggedright\arraybackslash}p{#1}}
\newcolumntype{C}[1]{>{\centering\arraybackslash}p{#1}}
\newcolumntype{R}[1]{>{\raggedleft\arraybackslash}p{#1}}

\ifCLASSOPTIONcompsoc
\usepackage[nocompress]{cite}
\else
\usepackage{cite}
\fi

\ifCLASSINFOpdf
\else
\fi

\begin{document}

	\title{
Deep Deformation Detail Synthesis \\ for Thin Shell Models
}
	
	\author{Lan Chen, Lin Gao\thanks{\IEEEauthorrefmark{1} Corresponding authors are Lin Gao (gaolin@ict.ac.cn) and Shibiao Xu (shibiao.xu@ia.ac.cn).}\IEEEauthorrefmark{1}, Jie Yang, Shibiao Xu\IEEEauthorrefmark{1}, Juntao Ye, Xiaopeng Zhang, Yu-Kun Lai 
		\IEEEcompsocitemizethanks{
			\IEEEcompsocthanksitem L. Chen is with Institute of Automation, Chinese Academy of Sciences, Beijing, China; and the School of Artificial Intelligence, University of Chinese Academy of Sciences, Beijing, China.\protect\\
			E-mail:chenlan2016@ia.ac.cn
			\IEEEcompsocthanksitem L. Gao and J. Yang are the Beijing Key Laboratory of Mobile Computing and Pervasive Device, Institute of Computing Technology, Chinese Academy of Sciences and also with the University of Chinese Academy of Sciences, Beijing, China.\protect\\
			E-mail:$\{$gaolin, yangjie01$\}$@ict.ac.cn
			\IEEEcompsocthanksitem S.B. Xu, J.T. Ye and X.P. Zhang are with Institute of Automation, Chinese Academy of Sciences, Beijing, China.\protect\\ 
			E-mail:$\{$shibiao.xu, juntao.ye, xiaopeng.zhang$\}$@ia.ac.cn   
			\IEEEcompsocthanksitem Y.-K. Lai is with Visual Computing Group, School of Computer Science and Informatics, Cardiff University, Wales, UK.\protect\\
			E-mail:LaiY4@cardiff.ac.uk
			}%
			\thanks{Manuscript received , 2020;}}
 
    \markboth{IEEE TRANSACTIONS ON VISUALIZATION AND COMPUTER GRAPHICS,~Vol.~xx, No.~xx, February~2021}%
    {Chen \MakeLowercase{\textit{et al.}}: Deep Deformation Detail Synthesis for Thin Shell Models}
	
	\IEEEtitleabstractindextext{%
		\begin{abstract}
			In %
			\gl{physics-based} cloth animation, rich folds and detailed wrinkles are achieved at the cost of expensive computational resources and huge labor tuning.
			Data-driven techniques \gl{make efforts to }reduce the computation significantly by utilizing a preprocessed database.
			One type of methods relies on human poses to synthesize fitted garments, but these methods cannot be applied to general cloth animations.
			Another type of methods adds details to the coarse meshes, which does not have such restrictions.
			However, existing works usually utilize coordinate-based representations which cannot cope with large-scale deformation, and requires \gl{dense vertex} correspondences between coarse and fine meshes. Moreover, as such methods only add details, they require coarse meshes to be sufficiently close to fine meshes, which can be either impossible, or require unrealistic constraints to be applied when generating \cl{fine} meshes.
			To address these \gl{challenges}, we \gl{develop a temporally and spatially as-consistent-as-possible deformation} representation (named TS-ACAP) and \gl{design} a DeformTransformer network to learn the mapping from low-resolution meshes to ones with fine details.
			This TS-ACAP representation is designed to ensure both spatial and temporal consistency for sequential large-scale deformations from cloth animations.
            With this TS-ACAP representation, our DeformTransformer network first utilizes two mesh-based encoders to extract the coarse and fine features using shared convolutional kernels, respectively.
			To transduct the coarse features to the fine ones, we leverage the Transformer network that consists of frame-level attention mechanisms to ensure temporal coherence of the prediction. 
			Experimental results show that our method is able to produce reliable and realistic animations in various datasets at high frame rates: $10 \sim 35$ times faster than physics-based simulation, with superior detail synthesis abilities compared to existing methods.
		\end{abstract}
		
		\begin{IEEEkeywords}
			Cloth animation, deep learning, large scale deformation, temporal consistency.
	\end{IEEEkeywords}}

	\maketitle

	\IEEEdisplaynontitleabstractindextext
	
	\IEEEpeerreviewmaketitle

	\input{sec/samplebody-journals}

	\vspace{-10mm}
	\begin{IEEEbiography}[{\includegraphics[width=1in,height=1.25in,clip,keepaspectratio]{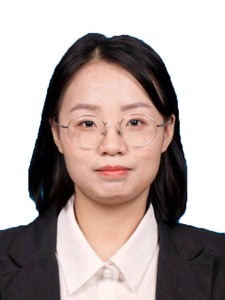}}]{Lan Chen}
		received her bachelor's degree in mathematics from China University of Petroleum - Beijing in 2016. She
		is currently a PhD student at Institute of Automation, Chinese Academy of Sciences. Her research interests include computer graphics and image processing. 
	\end{IEEEbiography}
	\vspace{-10mm}
	\begin{IEEEbiography}[{\includegraphics[width=1in,height=1.25in,clip,keepaspectratio]{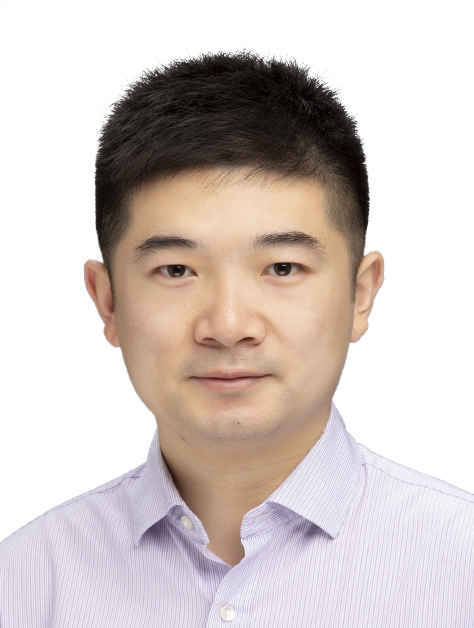}}]{Lin Gao}
		received the bachelor's degree in mathematics from Sichuan University and the PhD degree in computer science from Tsinghua University. He is currently an Associate Professor at the Institute of Computing Technology, Chinese Academy of Sciences. He has been awarded the Newton Advanced Fellowship from the Royal Society and the AG young researcher award. His research interests include computer graphics and geometric processing. 
	\end{IEEEbiography}
	\vspace{-10mm}
	\begin{IEEEbiography}[{\includegraphics[width=1in,height=1.25in,clip,keepaspectratio]{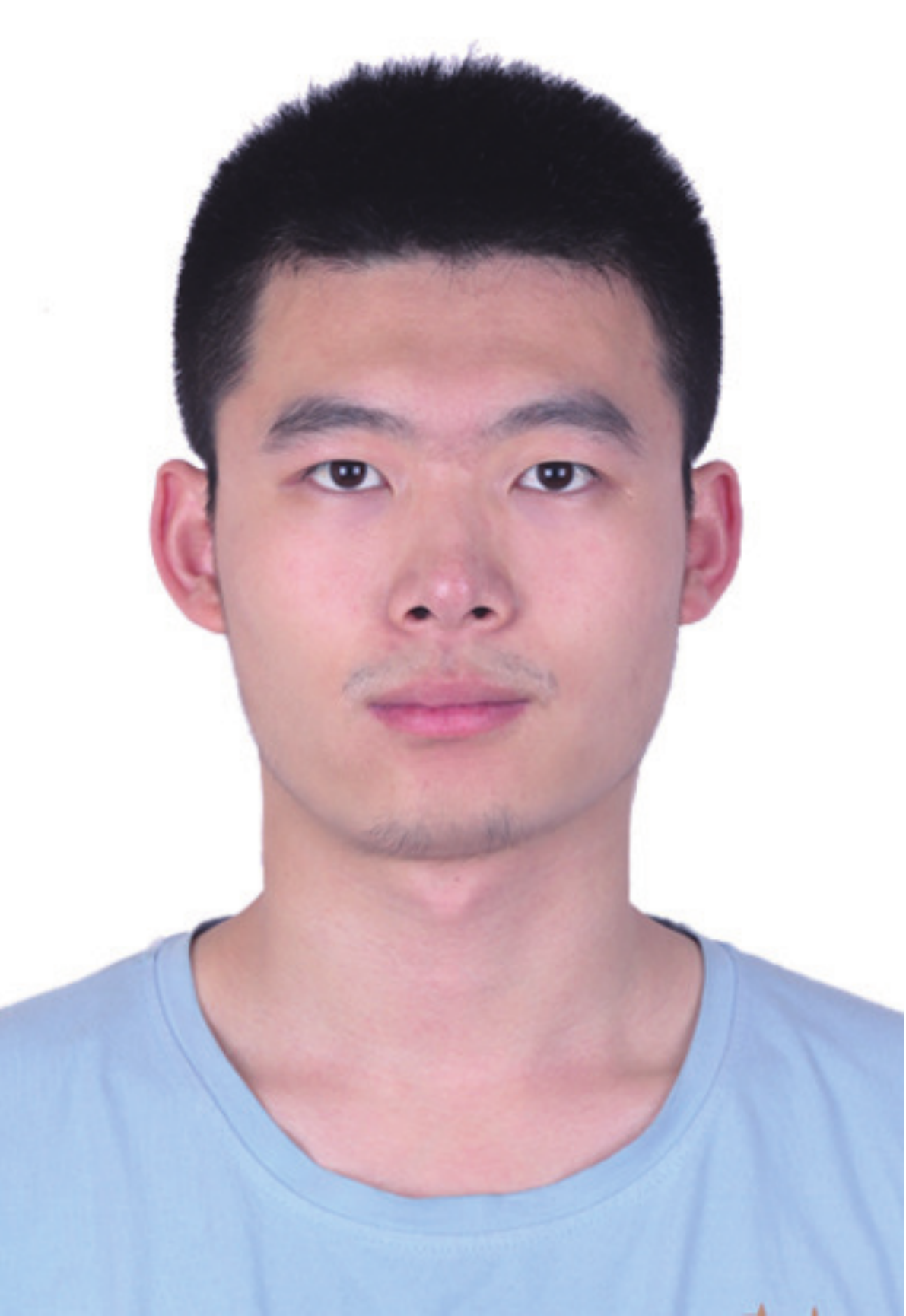}}]{Jie Yang}
		received the bachelor's degree in mathematics from Sichuan University in 2016. He is currently a PhD candidate in the Institute of Computing Technology, Chinese Academy of Sciences. His research interests include computer graphics and geometric processing.
	\end{IEEEbiography}
	\vspace{-10mm}
	\begin{IEEEbiography}[{\includegraphics[width=1in,height=1.25in,clip,keepaspectratio]{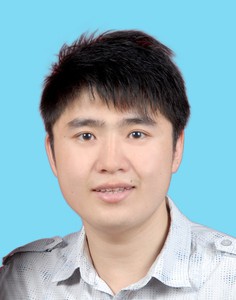}}]{Shibiao Xu}
    received the B.S. degrees in Information Engineering from Beijing University of Posts and Telecommunications in 2009, and the Ph.D. degree in Computer Science from Institute of Automation, Chinese Academy of Sciences in 2014. He is currently an Associate Professor with the National Laboratory of Pattern Recognition, Institute of Automation, Chinese Academy of Sciences. His current research interests include vision understanding and image-based three-dimensional reconstruction.
	\end{IEEEbiography}	
	\vspace{-10mm}
	\begin{IEEEbiography}[{\includegraphics[width=1in,height=1.25in,clip,keepaspectratio]{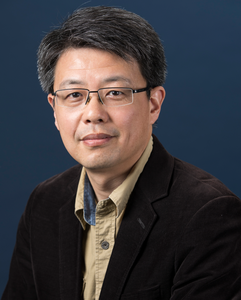}}]{Juntao Ye}
		was awarded his B.Eng from Harbin Engineering University in 1994, MSc from Institute of Computational Mathematics and Sci/Eng Computing, Chinese Academy of Sciences in 2000, and his PhD in Computer Science from University of Western Ontario, Canada in 2005. He is currently an associate professor at Institute of Automation, Chinese Academy of
		Sciences. His research interests include computer graphics and image processing, particularly simulation of cloth and fluid.
	\end{IEEEbiography}
	\vspace{-10mm}
	\begin{IEEEbiography}[{\includegraphics[width=1in,height=1.25in,clip,keepaspectratio]{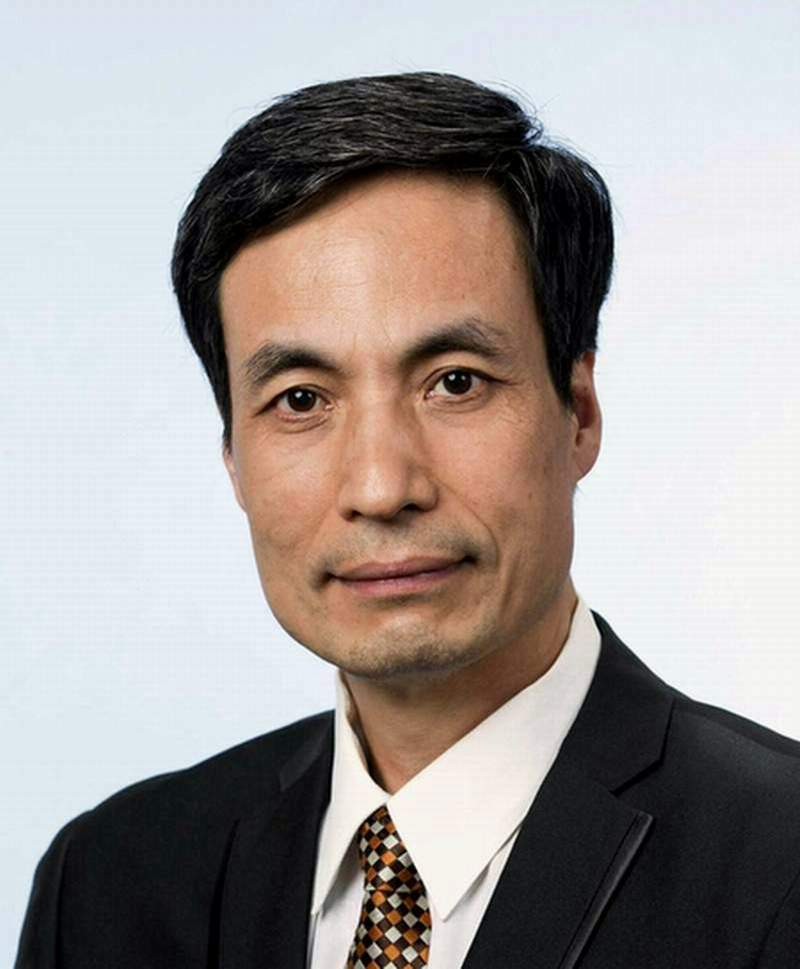}}]{Xiaopeng Zhang} received the PhD degree in computer science from Institute of Software,
	Chinese Academic of Sciences in 1999. He is a professor in National Laboratory of Pattern Recognition at Institute of Automation, Chinese Academy of Sciences. He received the National Scientific and Technological Progress Prize (second class) in 2004 and the Chinese Award of Excellent Patents in 2012. His main research interests include computer graphics and computer vision.
	\end{IEEEbiography}
	\vspace{-10mm}
	\begin{IEEEbiography}[{\includegraphics[width=1in,height=1.25in,clip,keepaspectratio]{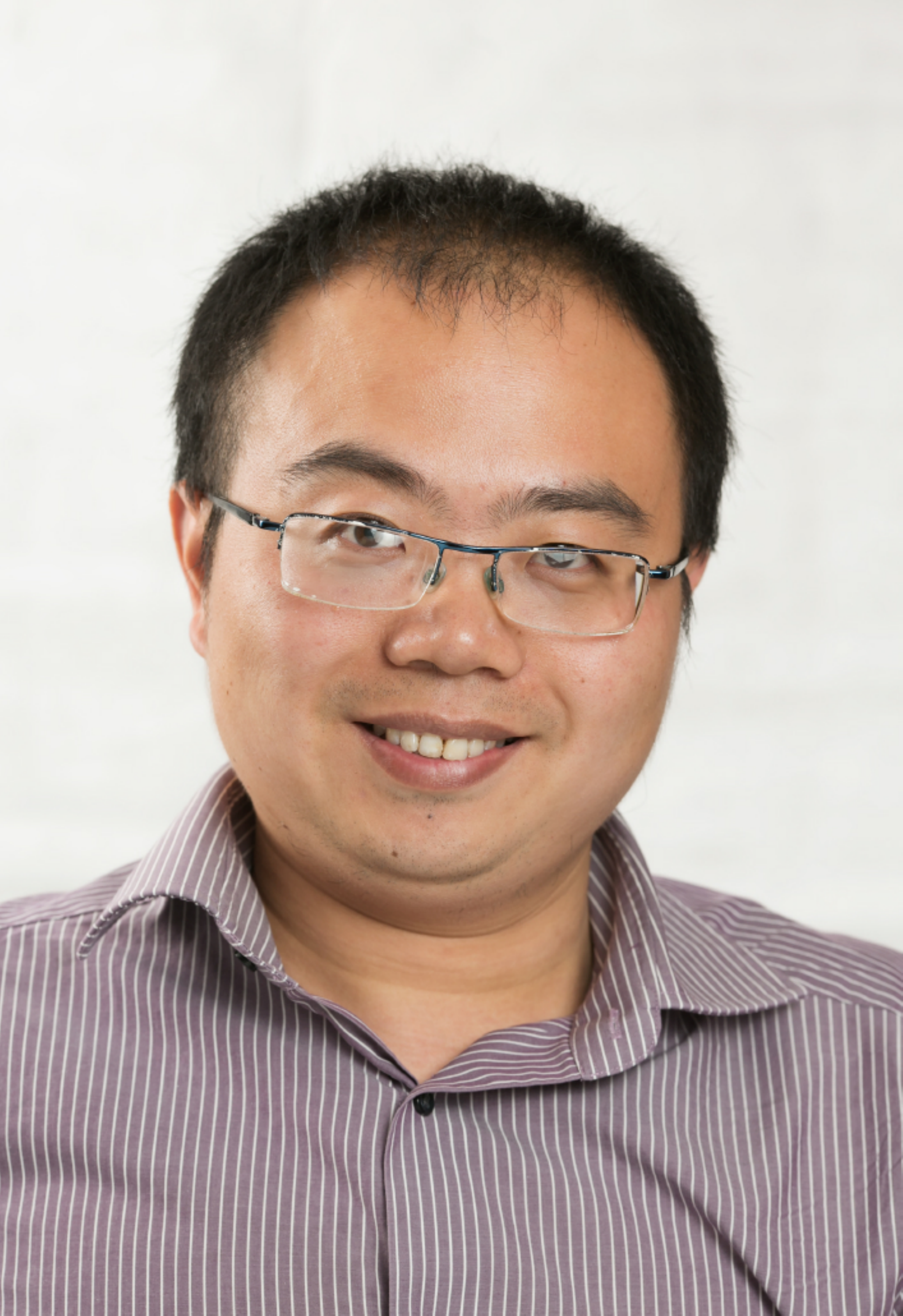}}]{Yu-Kun Lai}
		received his bachelor's degree and PhD degree in computer science from
		Tsinghua University in 2003 and 2008, respectively. He is currently a Professor in the School of Computer Science \& Informatics, Cardiff University. His research
		interests include computer graphics, geometry processing, image processing and computer vision. He is on the editorial boards of \emph{Computer Graphics Forum} and \emph{The Visual Computer}.
	\end{IEEEbiography}

\end{document}

%% file: sec/samplebody-journals.tex
\begin{figure}[t]
	\centering
	\begin{tabular}{ccc}
	\multicolumn{3}{c}{
	\includegraphics[width=1.0\linewidth]{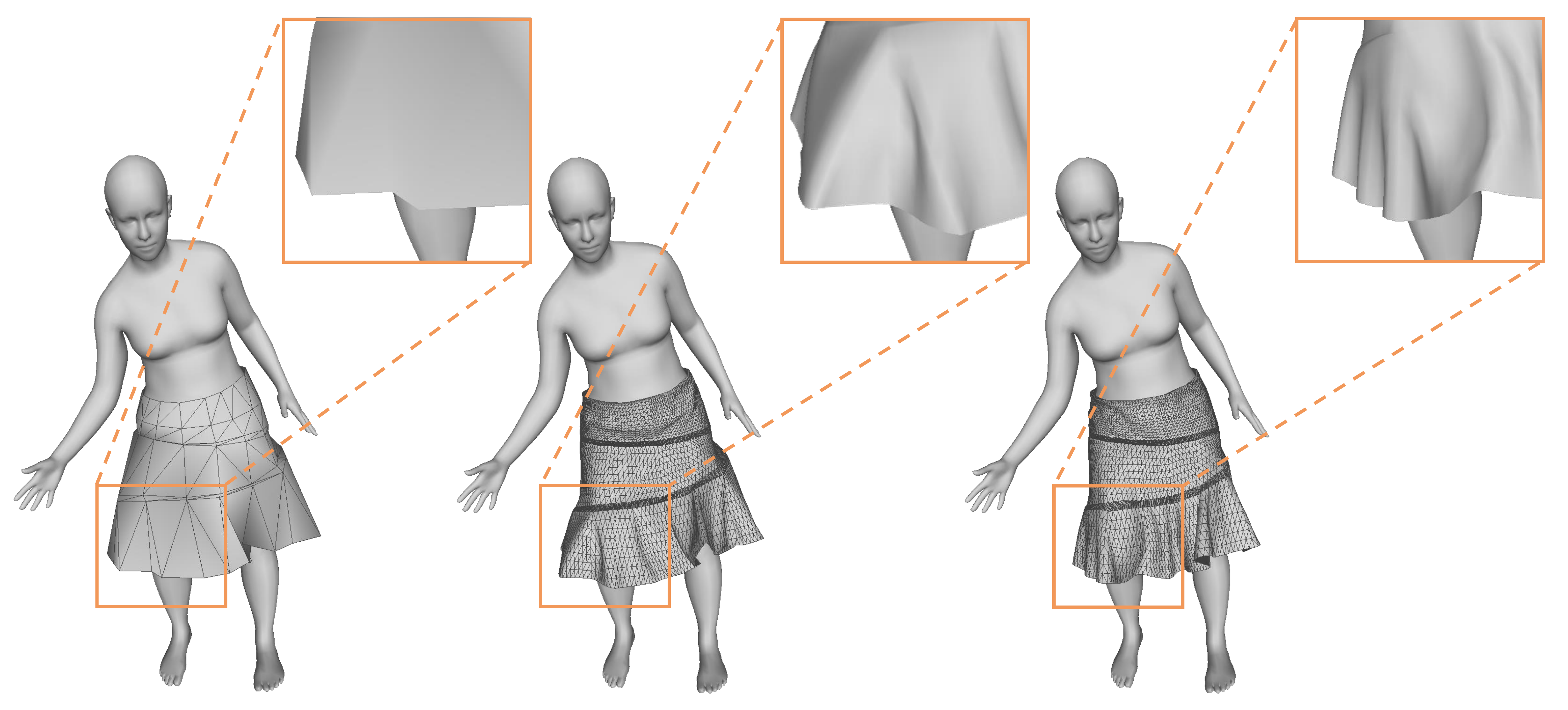}}  \\
	(a) coarse skirt & (b) tracked skirt & (c) fine skirt
	\end{tabular}
	\caption{\small \cl{One frame of \YL{skirt in different representations.} (a) \YL{coarse mesh} (207 triangles), (b) \YL{tracked mesh} (13,248 triangles) and (c) \YL{fine mesh} (13,248 triangles). \YL{Both coarse and fine meshes are obtained by simulating the skirt using a physics-based method \cl{\cite{Narain2012AAR}}. The tracked mesh is obtained with physics-based simulation involving additional constraints to track the coarse mesh.} The tracked mesh  exhibits stiff folds while the wrinkles in the fine simulated mesh are more realistic.}%
	}
	\label{fig:lrhrsim1} 
\end{figure}

\IEEEraisesectionheading{\section{Introduction}\label{sec:introduction}}
\IEEEPARstart{C}{reating} dynamic general clothes or garments on animated characters has been a long-standing problem in computer graphics (CG).
In the CG industry, physics-based simulations (PBS) are used to achieve realistic and detailed folding patterns for garment animations. 
However, it is time-consuming and requires expertise to synthesize fine geometric details since high-resolution meshes with tens of thousands or more vertices are often required.
For example, 10 seconds are required for physics-based simulation of a frame for detailed skirt animation shown in Fig.~\ref{fig:lrhrsim1}.
Not surprisingly, garment animation remains a bottleneck in many applications.
Recently, data-driven methods provide alternative solutions to fast and effective wrinkling behaviors for garments.
Depending on human body poses, some data-driven methods~\cite{wang10example,Feng2010transfer,deAguiar10Stable,santesteban2019learning, wang2019learning} are capable of generating tight cloth animations successfully.
Unfortunately, they are not suitable for loose garments, such as skirts, since the deformation of wrinkles cannot be defined by a static mapping from a character’s pose.
Instead of human poses, wrinkle augmentation on coarse simulations provides another alternative. 
It utilizes coarse simulations with fast speed to cover a high-level deformation and leverages learning-based methods to add realistic wrinkles.
Previous methods~\cite{kavan11physics,zurdo2013wrinkles,chen2018synthesizing} commonly require dense correspondences between coarse and fine meshes, so that local details can be added without affecting global deformation. 
\YL{Such methods also require coarse meshes to be sufficiently close to fine meshes, as they only add details to coarse meshes.}
To maintain the correspondences for training data and ensure closeness between coarse and fine meshes, weak-form constraints such as various test functions~\cite{kavan11physics,zurdo2013wrinkles,chen2018synthesizing} are applied to make fine meshes track the coarse meshes, 
\YL{but as a result, the obtained high-resolution meshes do not fully follow physical behavior, leading to animations that lack realism. An example is shown in Fig.~\ref{fig:lrhrsim1} where the tracked skirt (b) loses a large amount of wrinkles which should appear when simulating on fine meshes (c).}

Without requiring the constraints between coarse and fine meshes, we propose 
\gl{the DeformTransformer network
to synthesize detailed thin shell animations from coarse ones, based on deformation transfer.}
This is inspired by the similarity observed between pairs of coarse and fine meshes generated by PBS. %
Although the positions of vertices from two meshes are not aligned, the overall deformation is similar, so it is possible to predict fine-scale deformation with coarse simulation results.
Most previous works~\cite{kavan11physics,zurdo2013wrinkles,chen2018synthesizing} use explicit vertex coordinates to represent 3D meshes, which are sensitive to translations and rotations,
so they require good alignments between low- and high-resolution meshes. 
In our work, we regard the cloth animations as non-rigid deformation and propose a novel representation for mesh sequences, called TS-ACAP (Temporal and Spatial As-Consistent-As-Possible) representation. 
TS-ACAP is a local deformation representation, capable of representing and solving large-scale deformation problems, while maintaining the details of meshes.
Compared to the original ACAP representation~\cite{gao2019sparse}, TS-ACAP is fundamentally designed to ensure the temporal consistency of the extracted feature sequences, \YL{and meanwhile} it can maintain the original features of ACAP \YL{to cope with large-scale deformations}.
With \YL{TS-ACAP} representations for both coarse and fine meshes, we leverage a sequence transduction network to map the deformation from coarse to fine level to assure the temporal coherence of generated sequences.
Unlike existing works using recurrent neural networks (RNN)~\cite{santesteban2019learning}, we utilize the Transformer network~\cite{vaswani2017attention}, an architecture consisting of frame-level attention mechanisms for our mesh sequence transduction task.
It is based entirely on attention without recursion modules so can be trained significantly faster than architectures based on recurrent %
layers.
With \YL{temporally consistent features and the Transformer network, \YL{our method achieves} stable general cloth synthesis with fine details in an efficient manner.}

In summary, the main contributions of our work are as follows:
\begin{itemize}
	\item \YL{We propose a novel framework for the synthesis of cloth dynamics, by learning temporally consistent deformation from low-resolution meshes to high-resolution meshes \gl{with realistic dynamic}, which is $10 \sim 35$ times faster than PBS \cite{Narain2012AAR}.}
	\item \YL{To achieve this, we propose a \cl{temporally and spatially as-consistent-as-possible deformation representation (TS-ACAP)} to represent the cloth mesh sequences. It is able to deal with large-scale deformation, essential for mapping between coarse and fine meshes, while ensuring temporal coherence.} 
    \item \gl{Based on the TS-ACAP, We further design an effective neural network architecture (named DeformTransformer) by improving Transformer network, which successfully enables high-quality synthesis of dynamic wrinkles with rich details on thin shells and maintains temporal consistency on the generated high-resolution mesh sequences.}

\end{itemize}

We qualitatively and quantitatively evaluate our method for various cloth types (T-shirts, pants, skirts, square and disk tablecloth) with different motion sequences. 
In Sec.~\ref{sec:related_work}, we review the work most related to ours. We then give the detailed description of our method in Sec.~\ref{sec:approach}. 
Implementation details are presented in Sec.~\ref{sec:implementation}. We present experimental results, including extensive
comparisons with state-of-the-art methods in Sec.~\ref{sec:results}, and finally, we draw conclusions and \YL{discuss future work} in Sec.~\ref{sec:conclusion}.

\section{Related work} \label{sec:related_work}
\subsection{Cloth Animation}
Physics-based techniques for realistic cloth simulation have been widely studied in computer graphics, \YL{using methods such as} implicit Euler integrator \cite{BW98,Harmon09asynchronous}, iterative optimization \cite{terzopoulos87elastically,bridson03wrinkles,Grinspun03shell}, collision detection and response \cite{provot97collision,volino95collision}, etc. 
\YL{Although such techniques can generate realistic cloth dynamics, }they are time consuming for detailed cloth synthesis, and the robustness and efficiency of simulation systems are also of concern.
\YL{To address these, alternative methods have been developed to generate} the dynamic details of cloth animation via adaptive techniques \cite{lee2010multi,muller2010wrinkle,Narain2012AAR}, data-driven approaches \cite{deAguiar10Stable, Guan12DRAPE, wang10example, kavan11physics,zurdo2013wrinkles} and deep learning-based methods \cite{chen2018synthesizing,gundogdu2018garnet,laehner2018deepwrinkles,zhang2020deep}, etc.

 Adaptive techniques \cite{lee2010multi, muller2010wrinkle} usually simulate a coarse model by simplifying the smooth regions and \YL{applying interpolation} to reconstruct the wrinkles, \YL{taking normal or tangential degrees of freedom into consideration.}  
Different from simulating a reduced model with postprocessing detail augmentation, Narain {\itshape et al.} \cite{Narain2012AAR} directly generate dynamic meshes in \YL{the} simulation phase through adaptive remeshing, at the expense of increasing \YL{computation time}.  

Data-driven methods have drawn much attention since they offer faster cloth animations than physical models.
With \YL{a} constructed database of \YL{high-resolution} meshes, researchers have proposed many techniques depending on the motions of human bodies with linear conditional models\cite{deAguiar10Stable, Guan12DRAPE} or secondary motion graphs \cite{Kim2013near, Kim2008drivenshape}.
However, these methods are limited to tight garments and not suitable for skirts or cloth with more freedom.
An alternative line \YL{of research} is to augment details on coarse simulations \YL{by exploiting knowledge from a} database of paired meshes, to generalize the performance to complicated testing scenes.
In this line, in addition to wrinkle synthesis methods \YL{based on} bone clusters \cite{Feng2010transfer} or human poses \cite{wang10example} for fitted clothes, there are some approaches \YL{that investigate how to} learn a mapping from a coarse garment shape to a detailed one for general \YL{cases} of free-flowing cloth simulation.
Kavan {\itshape et al.} \cite{kavan11physics} present linear upsampling operators to \YL{efficiently} augment \YL{medium-scale} details on coarse meshes.
Zurdo {\itshape et al.} \cite{zurdo2013wrinkles} define wrinkles as local displacements and use \YL{an} example-based algorithm to enhance low-resolution simulations.
\YL{Their approaches mean the} high-resolution cloth \YL{is} required to track \YL{the} low-resolution cloth, \YL{and thus cannot} exhibit full high-resolution dynamics.

Recently deep learning-based methods have been successfully applied for 3D animations of human \YL{faces}~\cite{cao2016real, jiang20183d}, hair \cite{zhang2018modeling, yang2019dynamic} and garments \cite{liu2019neuroskinning, wang2019learning}.
As for garment synthesis, some approaches \cite{laehner2018deepwrinkles, santesteban2019learning, patel2020tailornet} are proposed to utilize a two-stream strategy consisting of global garment fit and local \YL{wrinkle} enhancement.
L{\" a}hner {\itshape et al.} \cite{laehner2018deepwrinkles} present DeepWrinkles, \YL{which recovers} the global deformation from \YL{a} 3D scan system and \YL{uses a} conditional \YL{generative adversarial network} to enhance a low-resolution normal map.
Zhang {\itshape et al.} \cite{zhang2020deep} further generalize the augmentation method with normal maps to complex garment types as well as various motion sequences.
\YL{These approaches add wrinkles on normal maps \YL{rather than geometry}, and thus their effectiveness is restricted to adding fine-scale visual details, not large-scale dynamics.}
Based on \YL{the} skinning representation, some algorithms \cite{gundogdu2018garnet,santesteban2019learning} use neural networks to generalize garment synthesis algorithms to multiple body shapes. 
\YL{In addition, other works are} devoted to \YL{generalizing neural networks} to various cloth styles \cite{patel2020tailornet} or cloth materials \cite{wang2019learning}.
Despite tight garments dressed on characters, some deep learning-based methods \cite{chen2018synthesizing, oh2018hierarchical} are %
\YL{demonstrated to work for cloth animation with higher degrees} of freedom.
Chen {\itshape et al.} \cite{laehner2018deepwrinkles} represent coarse and fine meshes via geometry images and use \YL{a} super-resolution network to learn the mapping.
Oh {\itshape et al.} \cite{oh2018hierarchical} propose a multi-resolution cloth representation with \YL{fully} connected networks to add details hierarchically.
Since the \YL{free-flowing cloth dynamics are harder for networks to learn} than tight garments, the results of these methods have not reached the realism of PBS. \YL{Our method based on a novel deformation representation and network architecture has superior capabilities of learning the mapping from coarse and fine meshes, generating realistic cloth dynamics, while being much faster than PBS methods.}
 
 \begin{figure*}[ht]
 	\centering
 	\includegraphics[width=1.0\linewidth, trim=20 250 20 50,clip]{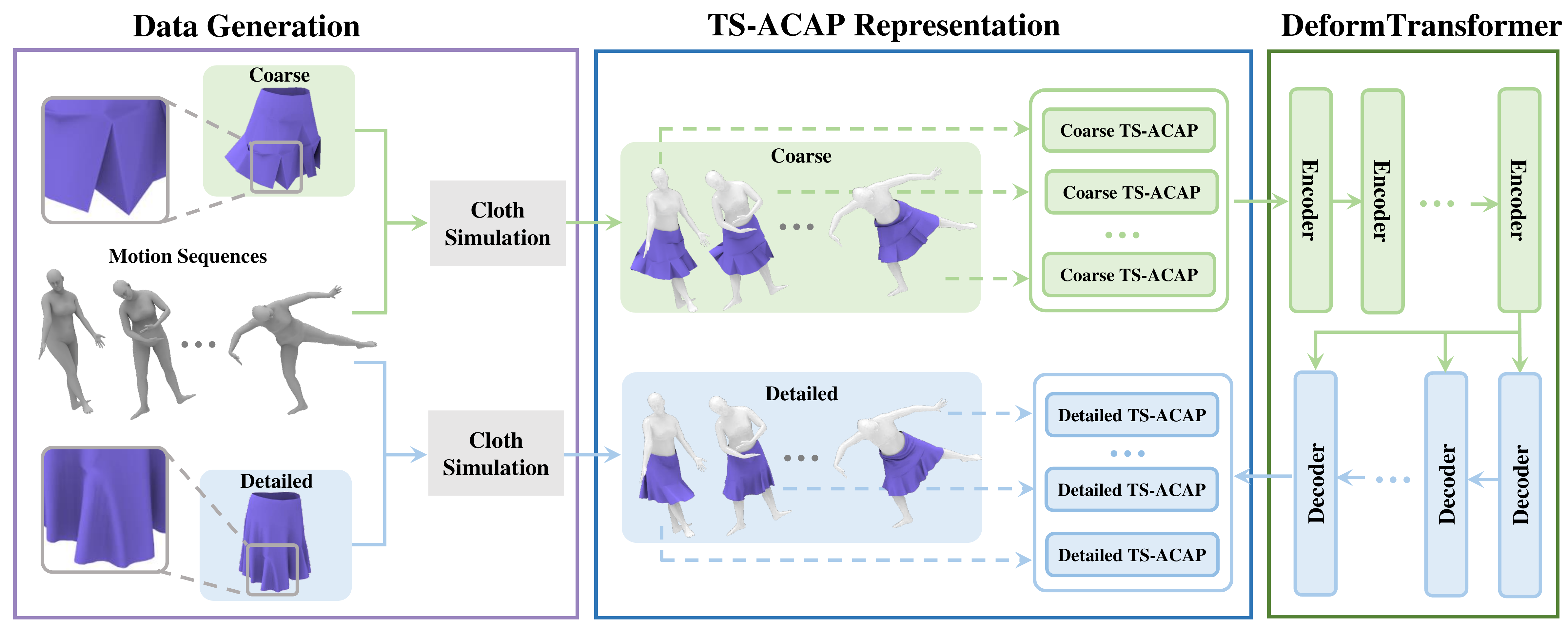} 
 	\caption{\small The overall architecture of our detail synthesis network. At data preparation stage, we generate low- and high-resolution \gl{thin shell} animations via coarse and fine \gl{meshes} and various motion sequences.
 	 Then we encode the coarse meshes and the detailed meshes to a deformation representation TS-ACAP, respectively.
 	\YL{Our algorithm then} learns to map the coarse features to fine features %
 	\YL{by designing a DeformTransformer network that consists of temporal-aware encoders and decoders, and finally reconstructs the detailed animations.}
 	}
 	\label{fig:pipeline}
 \end{figure*}

\subsection{Representation for 3D Meshes}
Unlike 2D images with regular grid of pixels, \YL{3D meshes have irregular connectivity which makes learning more difficult. To address this, existing deep learning based methods turn 3D meshes to a wide range of representations to facilitate processing~\cite{xiao2020survey},} such as voxels, images \YL{(such as depth images and multi-view images)}, point clouds, meshes, etc.
\YL{The volumetric representation has a regular structure, but it} often suffers from \YL{the problem of extremely high space and time consumption.}
Thus Wang {\itshape et al.} \cite{wang2017cnn} propose an octree-based convolutional neural network and encode the voxels sparsely. 
Image-based representations including \YL{depth images} \cite{eigen2014depth,gupta2014learning} and multi-view images \cite{Su2015mvcnn,li20193d} are proposed to encode 3D models in a 2D domain. 
It is unavoidable that both volumetric and image-based representations lose some geometric details.
Alternatively, geometry images are used in \cite{sinha2016deep,Sinha2017surfnet,chen2018synthesizing} for mesh classification or generation\YL{, which are obtained through cutting a 3D mesh to a topological disk, parameterizing it to a rectangular domain and regularly sampling the 3D coordinates in the 2D domain~\cite{gu2002geometry}.}
\YL{However, this representation} may suffer from parameterization distortion and seam line problems.

Instead of representing 3D meshes into other formats, recently there are methods \cite{tan2017autoencoder, tan2017variational, hanocka2019meshcnn} applying neural networks directly to triangle meshes with various features.
Gao {\itshape et al.} \cite{gao2016efficient} propose a deformation-based representation, called the rotation-invariant mesh difference (RIMD) which is translation and rotation invariant.
Based on the RIMD feature, Tan {\itshape et al.} \cite{tan2017variational}  propose a fully connected variational autoencoder network to analyze and generate meshes.
Wu {\itshape et al.} \cite{wu2018alive} use the RIMD to generate
a 3D caricature model from a 2D caricature image. 
However, it is expensive to reconstruct vertex coordinates from the RIMD feature due to the requirement of solving a very complicated optimization.
Thus it is not suitable for fast mesh generation tasks.
A faster deformation representation based on an as-consistent-as-possible (ACAP) formulation \cite{gao2019sparse} is further used to reconstruct meshes \cite{tan2017autoencoder}, which is able to cope with large rotations and efficient for reconstruction.
Jiang {\itshape et al.} \cite{jiang2019disentangled} use ACAP to disentangle the identity and expression of 3D \YL{faces}. 
They further apply ACAP to learn and reconstruct 3D human body models using a coarse-to-fine pipeline \cite{jiang2020disentangled}. 
\YL{However, the ACAP feature is represented based on individual 3D meshes. When applied to a dynamic mesh sequence, it does not guarantee temporal consistency.}
We propose a \cl{temporally and spatially as-consistent-as-possible (TS-ACAP)} representation, to ensure both spatial and temporal consistency of mesh deformation.
Compared to ACAP, our TS-ACAP can also accelerate the computation of features thanks to the sequential constraints.  

\subsection{Sequence Generation with \YL{DNNs (Deep Neural Networks)}}
Temporal information is crucial for stable and \gl{vivid} sequence generation. Previously, recurrent neural networks (RNN) have been successfully applied in many sequence generation tasks \cite{mikolov2010recurrent, mikolov2011extensions}. However, it is difficult to train \YL{RNNs} to capture long-term dependencies since \YL{RNNs} suffer from the vanishing gradient problem \cite{bengio1994learning}. To deal with this problem, previous works proposed some variations of RNN, including long short-term memory (LSTM) \cite{hochreiter1997long} and gated recurrent unit (GRU) \cite{cho2014properties}. These variations of RNN rely on the gating mechanisms to control the flow of information, thus performing well in the tasks that require capturing long-term dependencies, such as speech recognition \cite{graves2013speech} and machine translation \cite{bahdanau2014neural, sutskever2014sequence}. Recently, based on attention mechanisms, the Transformer network \cite{vaswani2017attention} has been verified to outperform \YL{many typical sequential models} for long sequences. This structure is able to inject the global context information into each input. Based on Transformer, impressive results have been achieved in tasks with regard to audio, video and text, \eg speech synthesis \cite{li2019neural, okamoto2020transformer}, action recognition \cite{girdhar2019video} and machine translation \cite{vaswani2017attention}.
We utilize the Transformer network to learn the frame-level attention which improves the temporal stability of the generated animation sequences.

\section{Approach} \label{sec:approach}
With a simulated sequence of coarse meshes $\mathcal{C} = \{\mathcal{C}_1, \dots, \mathcal{C}_n\}$ as input, our goal is to produce a sequence of fine ones $\mathcal{D} = \{\mathcal{D}_1, \dots, \mathcal{D}_n\}$ which have similar non-rigid deformation as the PBS. Given two simulation sets of paired coarse and fine garments, we extract the TS-ACAP representations respectively, \YL{and} then use our proposed DeformTransformer network to learn the \YL{transform} \YL{from the low-resolution space to the high-resolution space}. \YL{As illustrated previously in Fig.~\ref{fig:lrhrsim1}, such a mapping involves deformations beyond adding fine details.}
Once the network is trained by the paired examples, a consistent and detailed animation $\mathcal{D}$ can be synthesized for each input sequence $\mathcal{C}$. 

\subsection{Overview}
The overall architecture of our detail synthesis network is illustrated in Fig. \ref{fig:pipeline}.
To synthesize realistic \gl{cloth animations}, we propose a method to simulate coarse meshes first and learn a \YL{temporally-coherent} mapping to the fine meshes.    
To realize our goal, we construct datasets including low- and high-resolution cloth animations, \eg coarse and fine garments dressed on a human body of various motion sequences. 
To efficiently extract localized features with temporal consistency, we propose a new deformation representation, called TS-ACAP (temporal \YL{and spatial} as-consistent-as-possible), which is able to cope with both large rotations and unstable sequences. It also has significant advantages: it is efficient to compute for \YL{mesh} sequences and its derivatives have closed form solutions.
Since the vertices of the fine models are typically more than ten thousand to simulate realistic wrinkles, it is hard to directly map the coarse features to the high-dimensional fine ones for the network.
Therefore, \YL{convolutional encoder networks are} 
applied to encode \YL{coarse and fine meshes in the TS-ACAP representation} to \YL{their latent spaces}, respectively.
The TS-ACAP generates local rotation and scaling/shearing parts on vertices, so we perform convolution \YL{operations} on vertices %
\YL{to learn to extract useful features using shared local convolutional kernels.}
With encoded feature sequences, a sequence transduction network is proposed to learn the mapping from coarse to fine TS-ACAP sequences.
Unlike existing works using recurrent neural networks \YL{(RNNs)}~\cite{santesteban2019learning}, we use the Transformer \cite{vaswani2017attention}, a sequence-to-sequence network architecture, based on frame-level attention mechanisms for our detail synthesis task, \YL{which is more efficient to learn and leads to superior results.}

\subsection{Deformation Representation}
\YL{As discussed before, large-scale deformations are essential to represent \gl{thin shell mode dynamics such as }cloth animations, because folding and wrinkle patterns during animation can often be complicated. Moreover, cloth animations are in the form of sequences, hence the temporal coherence is very important for the realistic. Using 3D coordinates directly cannot cope with large-scale deformations well, and existing deformation representations are generally designed for static meshes, and directly applying them to cloth animation sequences on a frame-by-frame basis does not take temporal consistency into account. }
To cope with this problem, we propose a mesh deformation feature with spatial-temporal consistency, called TS-ACAP, to represent the coarse and fine deformed shapes, which exploits the localized information effectively and reconstructs \YL{meshes} accurately.
Take \YL{coarse meshes} $\mathcal{C}$ for instance and \YL{fine meshes $\mathcal{D}$ are processed in the same way.} \YL{Assume that a sequence} of coarse meshes contains $n$ models with the same topology, each denoted as $\mathcal{C}_{t}$ \YL{($1\leq t \leq n$)}. 
\YL{A mesh with the same topology is chosen as the reference model, denoted as $\mathcal{C}_{0}$. For example, for garment animation, this can be the garment mesh worn by  a character in the T pose.}
$\mathbf{p}_{t,i} \in \mathbb{R}^{3}$ is the $i^{\rm th}$ vertex on
the $t^{\rm th}$ mesh.
To represent the local shape deformation, the deformation gradient $\mathbf{T}_{t,i} \in \mathbb{R}^{3 \times 3}$ can be obtained by minimizing the following energy:
\begin{equation}
	\mathop{\arg\min}_{\mathbf{T}_{t,i}} \ \  \mathop{\sum}_{j \in \mathcal{N}_i} c_{ij} \| (\mathbf{p}_{t,i} - \mathbf{p}_{t,j}) -  \mathbf{T}_{t,i} (\mathbf{p}_{0,i} - \mathbf{p}_{0,j}) \|_2^2  \label{con:computeDG}
\end{equation}
where $\mathcal{N}_i$ is the one-ring neighbors of the $i^{\rm th}$ vertex, and $c_{ij}$ is the cotangent weight $c_{ij} = \cot \alpha_{ij} + \cot \beta_{ij} $ \cite{sorkine2007rigid,levi2014smooth}, where $\alpha_{ij}$
and $\beta_{ij}$ are angles opposite to the edge connecting the $i^{\rm th}$ and $j^{\rm th}$ vertices.

The main drawback of the deformation gradient representation is that it cannot handle large-scale rotations, which often \YL{happen} in cloth animation. 
Using polar decomposition, the deformation gradient $\mathbf{T}_{t,i} $ can be decomposed into a rotation part and a scaling/shearing part $\mathbf{T}_{t,i} = \mathbf{R}_{t,i}\mathbf{S}_{t,i}$.
The scaling/shearing transformation $\mathbf{S}_{t,i}$ is uniquely defined, while the rotation $\mathbf{R}_{t,i}$ \YL{corresponds to infinite possible rotation angles (differed by multiples of $2\pi$, along with possible opposite orientation of the rotation axis)}. Typical formulation often constrain the rotation angle to be within $[0, \pi]$ which is unsuitable for smooth large-scale animations. 

In order to handle large-scale rotations, we first require the orientations of rotation axes and rotation angles of \YL{spatially} adjacent vertices \YL{on the same mesh} to be as consistent as possible. 
Especially for our sequence data, we further add constraints for adjacent frames to ensure the temporal consistency of the orientations of rotation axes and rotation angles on each vertex.

We first consider consistent orientation for axes.
\begin{flalign}\label{eqn:axis}
	\arg\max_{{o}_{t,i}} \sum_{(i,j) \in \mathcal{E} }  {o}_{t,i}{o}_{t,j} \cdot s(\boldsymbol{\omega}_{t,i} \cdot \boldsymbol{\omega}_{t,j}, \theta_{t,i}, \theta_{t,j}) \nonumber\\
	+ \sum_{i \in \mathcal{V} }  {o}_{t,i} \cdot s(\boldsymbol{\omega}_{t,i} \cdot \boldsymbol{\omega}_{t-1,i}, \theta_{t,i}, \theta_{t-1,i}) \nonumber\\
	{\rm s.t.} \quad
	{o}_{t,1} = 1, {o}_{t,i} = \pm 1 (i \neq 1) \quad  
\end{flalign}
where $t$ is the \YL{index} of \YL{the} frame, $\mathcal{E}$ is the edge set, and $\mathcal{V}$ is the vertex set. \YL{Denote by $(\boldsymbol{\omega}_{t,i}, \theta_{t,i})$ one possible choice for the rotation axis and rotation angle that match $\mathbf{R}_{t,i}$. $o_{t,i} \in \{+1, -1\}$ specifies whether the rotation axis is flipped ($o_{t,i} = 1$ if the rotation axis is unchanged, and $-1$ if its opposite is used instead). }\YL{The first term promotes spatial consistency while the second term promotes temporal consistency.} 
$s(\cdot)$ is a function measuring orientation consistency, which is defined as follows:
\begin{equation}
	s(\cdot)=\left\{
	\begin{aligned}
		0 & , & |\boldsymbol{\omega}_{t,i} \cdot \boldsymbol{\omega}_{t,j}|\leq\epsilon_1 \; {\rm or} \;
		\theta_{t,i}<\varepsilon_2 \; {\rm or}  \; \theta_{t,j}<\varepsilon_2 \\
		1 & , & {\rm Otherwise~if}~\boldsymbol{\omega}_{t,i} \cdot \boldsymbol{\omega}_{t,j}>\epsilon_1 \\
		-1 & , & {\rm Otherwise~if}~ \boldsymbol{\omega}_{t,i} \cdot \boldsymbol{\omega}_{t,j}<-\epsilon_1 \\
	\end{aligned}
	\right.
\end{equation}
\YL{The first case here is to ignore cases where the rotation angle is near zero, as the rotation axis is not well defined in such cases.}
As for rotation angles, \YL{we optimize the following}
\begin{flalign}\label{eqn:angle}
\arg\min_{r_{t,i}} &\sum_{(i,j) \in \mathcal{E} } \| (r_{t,i} \cdot 2\pi+{o}_{t,i}\theta_{t,i}) - (r_{t,j}\cdot 2\pi+{o}_{t,j}\theta_{t,j}) \|_2^{2} &\nonumber\\
+ &\sum_{i \in \mathcal{V} } \| (r_{t,i} \cdot 2\pi+{o}_{t,i}\theta_{t,i}) - (r_{t-1,i}\cdot 2\pi+{o}_{t,j}\theta_{t-1,i}) \|_2^{2} \nonumber\\ 
{\rm s.t.}& \quad r_{t,i} \in \mathbb{Z},~~r_{t,1} = 0.
\end{flalign}
where $r_{t,i} \in \mathbb{Z}$ specifies how many $2\pi$ rotations should be added to the rotation angle.
\YL{The two terms here promote spatial and temporal consistencies of rotation angles, respectively. 
These optimizations can be solved using integer programming, and we use the mixed integer solver CoMISo~\cite{comiso2009} which provides an efficient \gl{solver}. See~\cite{gao2019sparse} for more details.}
A similar process is used to compute the TS-ACAP representation of the fine meshes.

\cl{Compared to the ACAP representation, our TS-ACAP representation considers temporal constraints to represent nonlinear deformation for optimization of axes and angles, which is more suitable for consecutive large-scale deformation \YL{sequences}.
We compare ACAP~\cite{gao2019sparse} and our TS-ACAP using a simple example of a simulated disk-shaped cloth animation sequence. Once we obtain deformation representations of the meshes in the sequence, 
we interpolate two meshes, the initial state mesh and a randomly selected frame, using linear interpolation of \YL{shape representations}.
\YL{In Fig. \ref{fig:interpolation}, we demonstrate the interpolation results with ACAP representation, which shows that it cannot handle such challenging cases with complex large-scale deformations. In contrast, with our temporally and spatially as-consistent-as-possible optimization, our TS-ACAP representation is able to produce consistent interpolation results.}

\begin{figure}[ht]
	\centering
	\includegraphics[width=\linewidth]{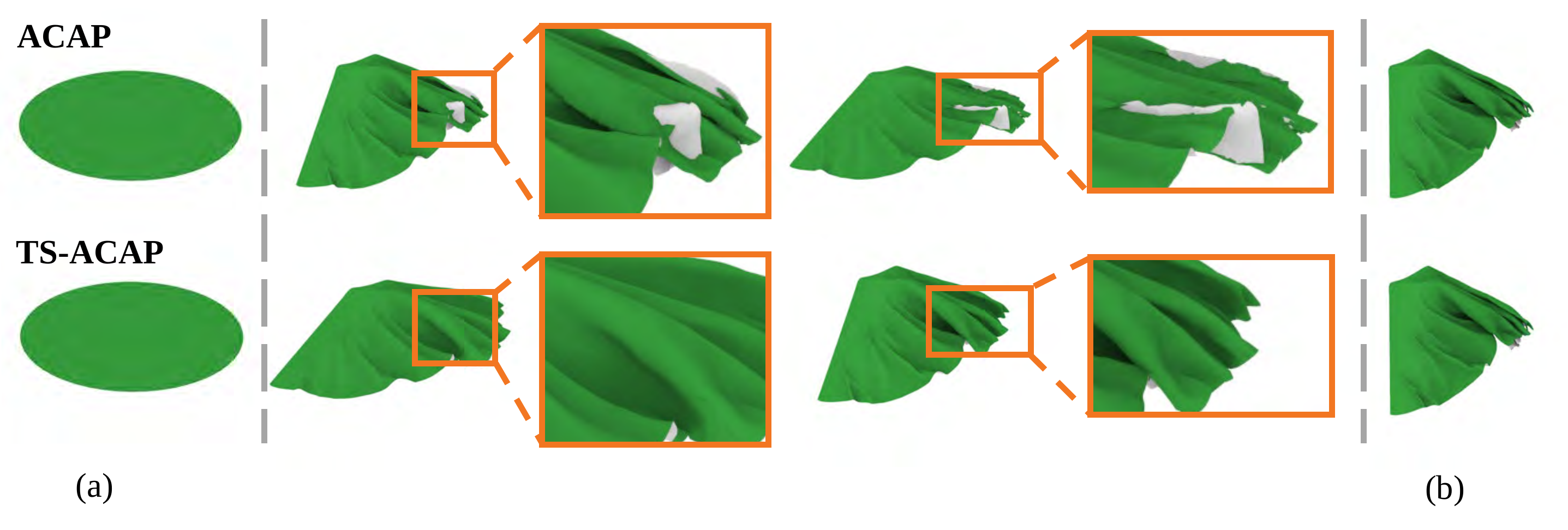}%
	\caption{\small Comparison of shape interpolation results with different deformation representations, ACAP and TS-ACAP.  %
	(a) and (b) are the source (t = 0) and target (t = 1) models with large-scale deformation to be interpolated. 
	The first row shows the interpolation results by ACAP, and the second row show the results with our TS-ACAP. 
	\gl{The interpolated models with ACAP feature are plausible in each frame while they are not consistent in the temporal domain.}
	}
	\label{fig:interpolation}
\end{figure}
}

\subsection{DeformTransformer Networks}
\begin{figure}[ht]
	\centering
	\includegraphics[width=0.48\linewidth]{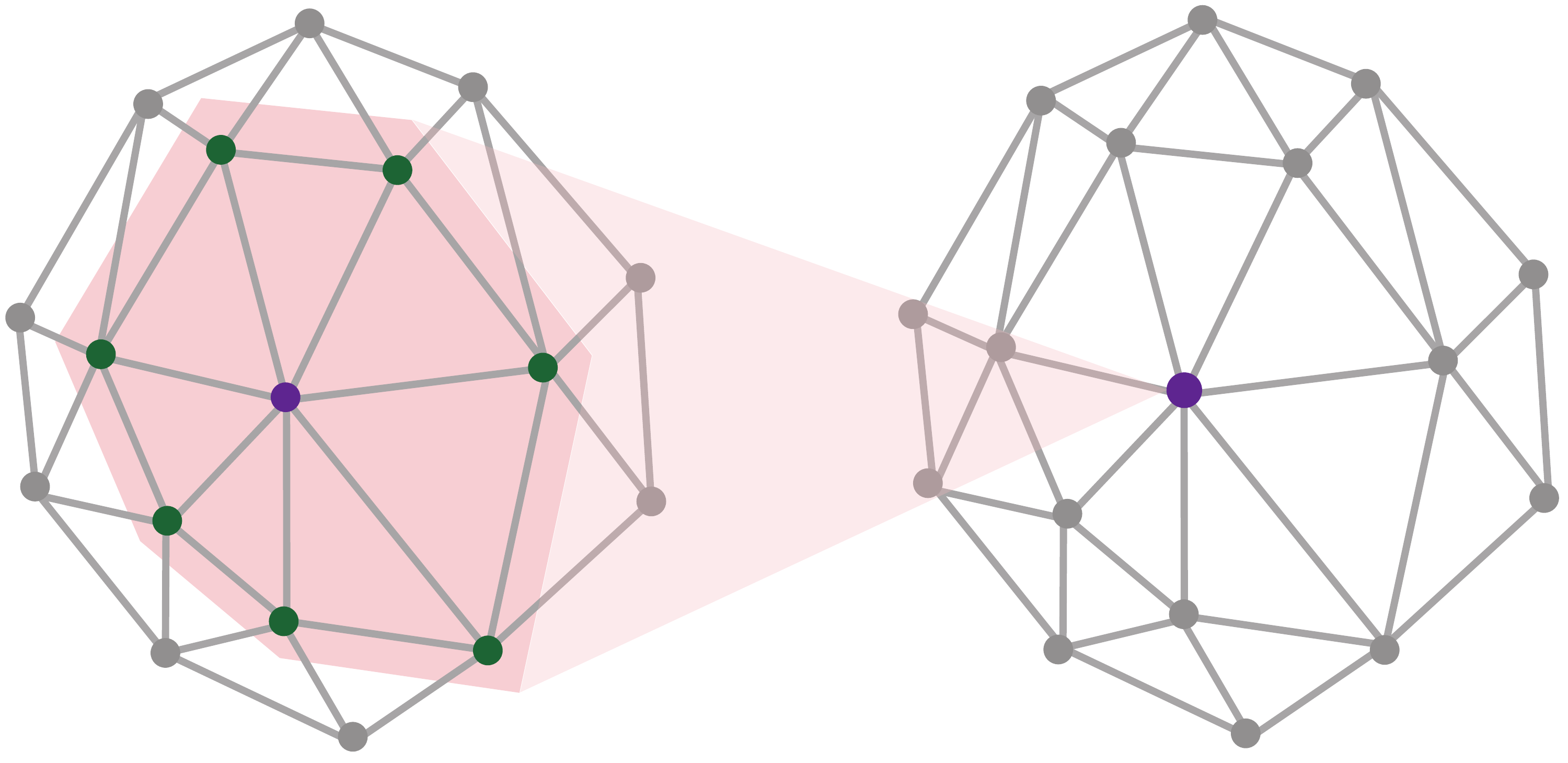} 
	\caption{\small Illustration of the convolutional operator on meshes. 
		The result of convolution for each vertex is obtained by a linear combination from the input in the 1-ring neighbors of the vertex, along with a bias.
	}
	\label{fig:pointconv}
\end{figure}

Unlike \cite{tan2017variational, wang2019learning} which use fully connected layers for mesh encoder, we perform convolutions \YL{on meshes to learn to extract useful features using compact shared convolutional kernels.} 
As illustrated in Fig. \ref{fig:pointconv}, we use a convolution operator on vertices  \cite{duvenaud2015convolutional, tan2017autoencoder} where the output at a vertex is obtained as a linear combination of input in its one-ring neighbors along with a bias. 
\YL{The input to our network is the TS-ACAP representation, which for the $i^{\rm th}$ vertex of the $t^{\rm th}$ mesh, we collect non-trivial coefficients from the rotation $\mathbf{R}_{t, i}$ and scaling/shearing $\mathbf{S}_{t,i}$, which forms a 9-dimensional feature vector (see~\cite{gao2019sparse} for more details). Denote by $\mathbf{f}_i^{(k-1)}$ and $\mathbf{f}_i^{k}$ the feature of the $i^{\rm th}$ vertex at the $(k-1)^{\rm th}$ and $k^{\rm th}$ layers, respectively. The convolution operator is defined as follows:
\begin{equation}
	\mathbf{f}_i^{(k)} =
	\mathbf{W}_{point}^{(k)} \cdot \mathbf{f}_{i}^{(k-1)} + 
	\mathbf{W}_{neighbor}^{(k)} \cdot \frac{1}{D_i} \mathop{\sum}_{j=1}^{D_i} \mathbf{f}_{n_{ij}}^{(k-1)}
	+ \mathbf{b}^{(k)} 
\end{equation}
where $\mathbf{W}_{point}^{(k)}$, $\mathbf{W}_{neighbor}^{(k)}$ and $\mathbf{b}^{(k)}$ are learnable parameters for the $k^{\rm th}$ convoluational layer, $D_i$ is the degree of the $i^{\rm th}$ vertex, $n_{ij}(1 \leq j \leq D_i )$ is the $j^{\rm th}$ neighbor of the $i^{\rm th}$ vertex.
}

\begin{figure}[ht]
	\centering
	\includegraphics[width=\linewidth, trim=0 50 0 150,clip]{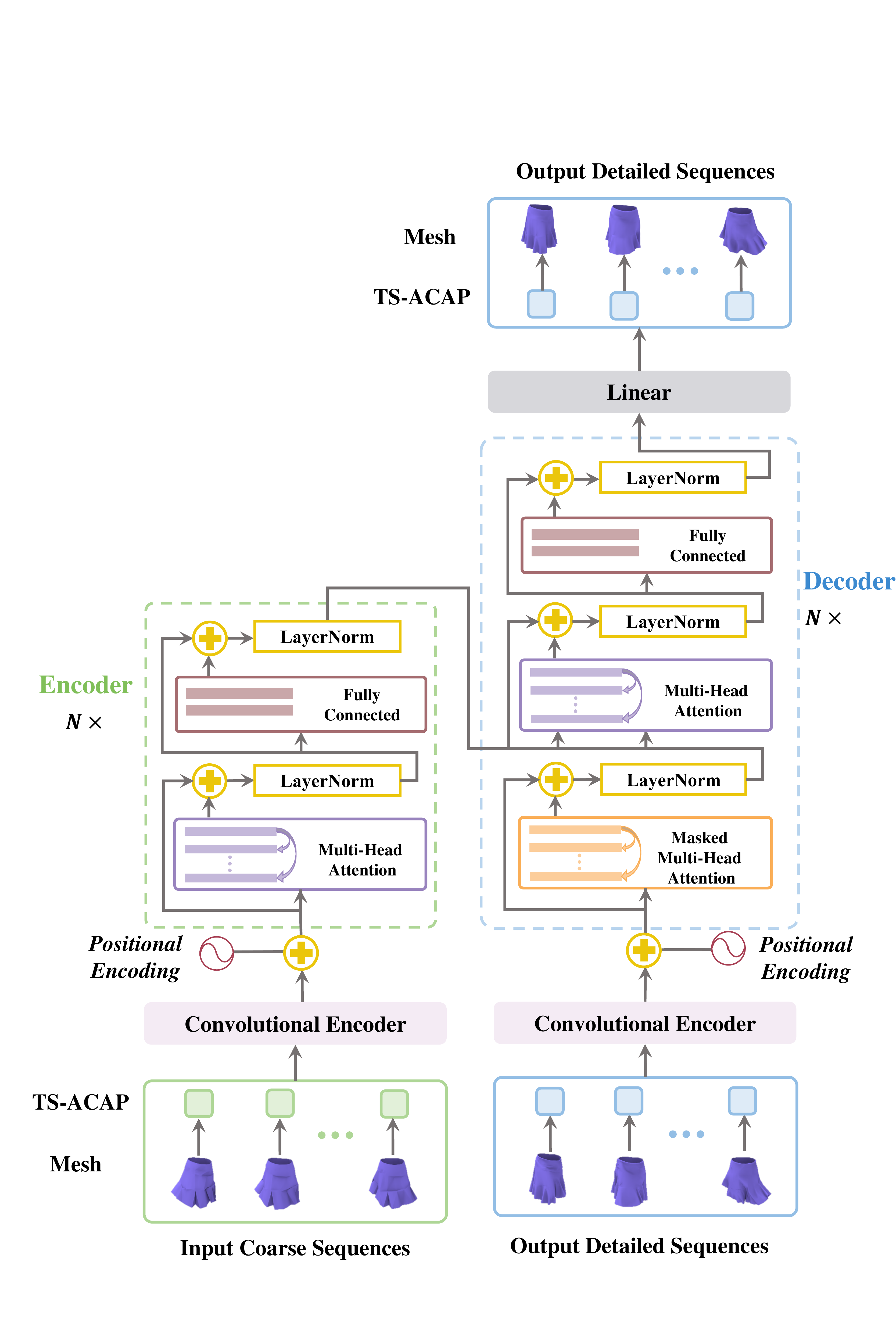} %
	\caption{\small The architecture of our DeformTransformer network.
		The coarse and fine mesh sequences are embedded into feature vectors using the TS-ACAP representation which \YL{is} defined \YL{at} each vertex as a 9-dimensional vector. 
		Then two convolutional \YL{encoders} map coarse and fine features to \YL{their latent spaces}, respectively.
		These latent vectors are fed into the DeformTransformer network, \cl{which consists of the encoder and decoder, each including a stack of $N=2$ identical blocks with 8-head attention,} to recover \YL{temporally-coherent} deformations.
		Notice that in \YL{the} training phase the input high-resolution TS-ACAP \YL{features are those from the ground truth}, 
		\YL{but during testing, these features are initialized to zeros, and once a new high-resolution frame is generated, its TS-ACAP feature is added.}
		With predicted feature vectors, realistic and stable cloth animations are generated.
	}
	\label{fig:Transformer}
\end{figure}

\begin{figure}[ht]
	\centering
	\includegraphics[width=0.4\linewidth, trim=18 33 18 3,clip]{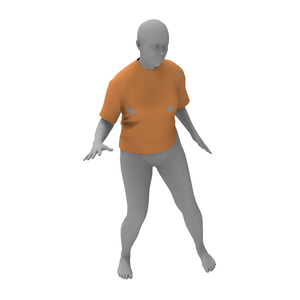} 
	\includegraphics[width=0.4\linewidth, trim=18 33 18 3,clip]{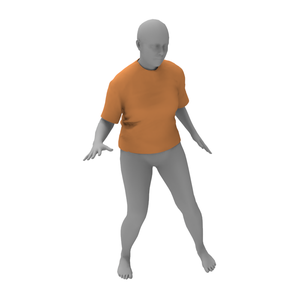} 
	\caption{\small For tight clothing, data-driven cloth deformations may suffer from apparent collisions with the body (left). We apply a simple postprocessing step to push 
	\YL{the collided} T-shirt vertices outside the body (right).
	}
	\label{fig:collisionrefinement}
\end{figure}
\begin{figure*}[ht]
	\centering
	\includegraphics[width=1.0\linewidth, trim=50 150 100 150,clip]{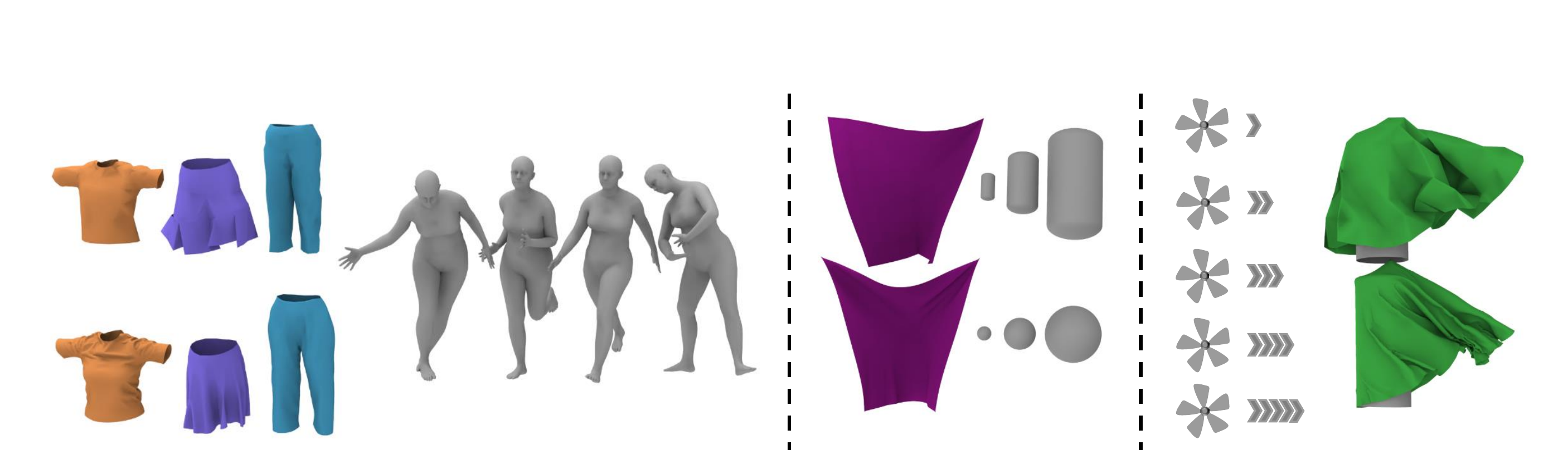} 
	\caption{\small 
		We test our algorithm on 5 datasets including TSHIRT, PANTS, SKIRT, SHEET and DISK.		 
		The former three are  garments (T-shirts, skirts, and pants) dressed on a template body and simulated with various motion sequences.
		The SHEET dataset is a square sheet interacting with various obstacles.
		The DISK dataset is a round tablecloth draping on a cylinder in the wind of various velocities. 
		Each cloth shape has a coarse resolution (top) and a fine resolution (bottom). 
	} 
	\label{fig:dataset}
\end{figure*}
Let $\mathcal{F}_\mathcal{C} = \{\mathbf{f}_{\mathcal{C}_1}, \dots, \mathbf{f}_{\mathcal{C}_n}\}$ be the sequence of coarse mesh features, and $\mathcal{F}_\mathcal{D} = \{\mathbf{f}_{\mathcal{D}_1}, \dots, \mathbf{f}_{\mathcal{D}_n}\}$ be its counterpart, the sequence of detailed mesh features.
To synthesize $\mathcal{F}_\mathcal{D}$ from $\mathcal{F}_\mathcal{C}$, the DeformTransformer framework is proposed to solve this sequence-to-sequence problem.
The DeformTransformer network consists of several stacked encoder-decoder layers, \YL{denoted} as $Enc(\cdot)$ and $Dec(\cdot)$. To take the order of the sequence into consideration, triangle positional embeddings \cite{vaswani2017attention} are injected into frames of $\mathcal{F}_\mathcal{C}$ and $\mathcal{F}_\mathcal{D}$, respectively.
The encoder takes coarse mesh features as input and encodes it to a \YL{temporally-dependent} hidden space.
It is composed of identical blocks \YL{each} with two sub-modules, one is the multi-head self-attention mechanism, the other is the frame-wise fully connected feed-forward network. 
We also employ a residual connection around these two sub-modules, followed \YL{by} the layer normalization.
The multi-head attention is able to build the dependence between any frames, thus ensuring that each input can consider global context of the whole sequence. Meanwhile, compared with other sequence models, this mechanism splits \YL{the} attention into several subspaces so that it can model the frame \YL{relationships} in multiple aspects.
With the encoded latent vector $Enc(\mathcal{F}_\mathcal{C})$, the decoder network attempts to reconstruct a sequence of fine mesh features.
The decoder has two parts: 
The first part takes fine mesh sequence $\mathcal{F}_\mathcal{D}$ as \YL{input} and 
encodes it similar to the encoder. 
\YL{Unlike the encoder, detailed meshes are generated sequentially, and when predicting frame $t$, it should not attend to subsequent frames (with the position after frame $t$). To achieve this, we utilize a masking process
for the self-attention module.} The second part performs multi-head attention over the output of the encoder, thus capturing the long-term dependence between coarse mesh features $\mathcal{F}_\mathcal{C}$ and fine mesh features $\mathcal{F}_\mathcal{D}$.
We train the Transformer network by minimizing the mean squared error between predicted detailed features and the ground-truth.
With predicted TS-ACAP feature vector, we reconstruct the vertex coordinates of \YL{the} target mesh\YL{, in the same way as reconstruction from ACAP features} (please refer to \cite{gao2019sparse} for details).   
Our training data is generated by PBS \YL{and is collision-free}.
Since human body \YL{(or other obstacles)} information is unseen in our algorithm, it does not guarantee the predicted cloth \YL{is free from any penetration}.
Especially for tight garment like T-shirts, it will be apparent if collision \YL{between the garment and human body} happens.
We use a fast refinement method \cite{wang2019learning} to push the cloth vertices colliding with the body outside \YL{while} preserving the local wrinkle details (see Fig.~\ref{fig:collisionrefinement}). 
For each vertex detected inside the body, we find its closest point over the body surface with normal and position.
Then the cloth mesh is deformed to update the vertices by minimizing the energy which penalizes the euclidean distance and Laplacian difference between the updated mesh and the initial one (please refer to \cite{wang2019learning} for details).
The collision solving process usually takes less than 3 iterations to converge to a collision-free state.

\section{Implementation}\label{sec:implementation}
We describe the details of the dataset construction and the network architecture in this section.

\textbf{\YL{Datasets}.}
To test our method, we construct 5 datasets, called TSHIRT, PANTS, SKIRT, SHEET and DISK respectively.
The former three datasets are different types of garments, \ie, T-shirts, skirts and pants worn on human bodies.
Each type of garment \YL{is represented by both low-resolution and  high-resolution meshes}, \YL{containing} 246 and 14,190 vertices for the T-shirts, 219 and 12,336 vertices for the skirts, 200 and 11,967 vertices for the pants.
Garments of the same type and resolution are simulated from a template mesh, which means \YL{such meshes obtained  through cloth animations have the same number of vertices and the same connectivity}.
These garments are dressed on animated characters, which are obtained via driving a body \YL{in the SMPL (Skinned Multi-Person Linear)  model} \cite{loper2015smpl} with publicly available motion capture data from CMU \cite{hodgins2015cmu}.
Since the motion data is captured, there are some \YL{self-collisions} or long repeated sequences.  
\YL{After removing poor quality data}, we select various motions, such as dancing, walking, running, jumping etc., including 20 sequences (\YL{9031, 6134, 7680 frames in total} for TSHIRT, PANTS and SKIRT respectively).
In these motions, 18 sequences are randomly selected for training and the remaining 2 sequences for testing.
The SHEET dataset consists of a pole or a sphere of three different sizes crashing to a piece of \YL{cloth sheet}.
The coarse mesh has 81 vertices and the fine mesh has 4,225 vertices.
There are \YL{4,000} frames in the SHEET dataset, in which 3200 frames for training and \YL{the remaining} 800 frames for testing.
We construct the DISK dataset by draping a round tablecloth to a cylinder in the wind, with 148 and 7,729 vertices for coarse and fine meshes respectively.
We adjust the velocity of the wind to get various animation sequences, in which 1600 frames for training and 400 frames for testing. 

\begin{table*}[ht]
	\renewcommand\arraystretch{1.5}
	\caption{ Statistics and timing (sec/\YL{frame}) of the testing examples including five types of \YL{thin shell animations}.
	}
	\label{table:runtime}
	\centering
	\begin{tabular}{cccccccccc}
		\toprule[1.2pt] 
		Benchmark & \#verts & \#verts & PBS & ours  & speedup & \multicolumn{4}{c}{our components}              \\ \cline{7-10} 
		& LR      & HR      & HR    &       &         & coarse & TS-ACAP & synthesizing & refinement \\
		&         &         &         &       &         & sim.   & extraction & (GPU)        &            \\  \hline \hline
		TSHIRT    & 246     & 14,190   &  8.72    & 0.867 & \textbf{10}     & 0.73   & 0.11      &  0.012 & 0.015\\
		PANTS     & 200     & 11,967   & 10.92  &0.904   & \textbf{12}      & 0.80   & 0.078 & 0.013  & 0.013\\
		SKIRT     & 127     & 6,812 & 6.84 & 0.207  & \textbf{33} & 0.081 & 0.10 & 0.014 & 0.012 \\  
		SHEET     & 81      & 4,225    & 2.48    & 0.157  & \textbf{16} & 0.035  & 0.10 &   0.011      &  0.011 \\  
		DISK      & 148     & 7,729    & 4.93 & 0.139 & \textbf{35}     & 0.078  & 0.041 &   0.012      &   0.008 \\  
		\bottomrule[1.2pt]
	\end{tabular}
\end{table*} 
To prepare the above datasets, we generate both \YL{low-resolution (LR)} and \YL{high-resolution (HR)} cloth \YL{animations} by PBS.
The initial state of the HR mesh is obtained by applying the Loop subdivision scheme \cite{Thesis:Loop} to the coarse mesh and waiting for several seconds till stable.
Previous works \cite{kavan11physics, zurdo2013wrinkles, chen2018synthesizing} usually constrain the high-resolution meshes by various tracking mechanisms to ensure that the coarse cloth \YL{can be seen as} a low-resolution version of the fine cloth during the complete animation sequences.
However, fine-scale wrinkle dynamics cannot be captured by this model, as wrinkles are defined quasistatically and limited to a \YL{constrained} subspace.
Thus we \YL{instead perform} PBS for the two resolution meshes \emph{separately}, without any constraints between them.
We use a cloth simulation engine called ARCSim \cite{Narain2012AAR} to produce all animation sequences of low- and high-resolution meshes with the same parameter setting. 
In our experiment, we choose the Gray Interlock from a library of measured cloth materials \cite{Wang2011DEM} as the material parameters for ARCSim simulation.
Specially for garments interacting with characters, to ensure collision-free, we manually put the coarse and fine garments on a template human body (in the T pose) and run the simulation to let the \YL{clothing} relax. To this end, we define the initial state for all subsequent simulations.
We interpolate 15 frames between the T pose and the initial pose of each motion sequence, before applying the motion sequence, which is smoothed using a convolution operation.

\begin{figure}[ht]
	\centering
	\subfloat{
		\includegraphics[width=0.5\linewidth]{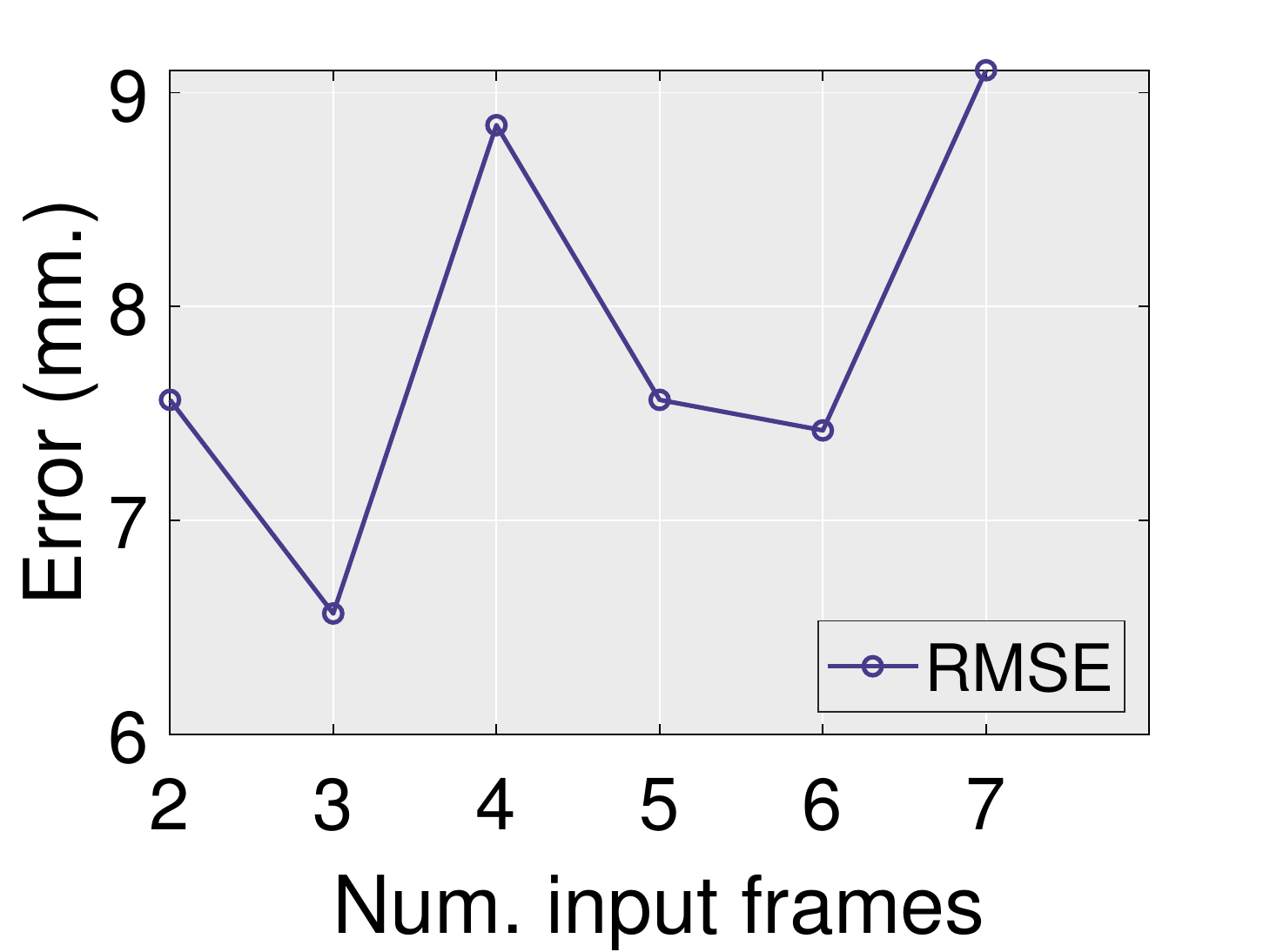} 
	}
	\subfloat{
		\includegraphics[width=0.5\linewidth]{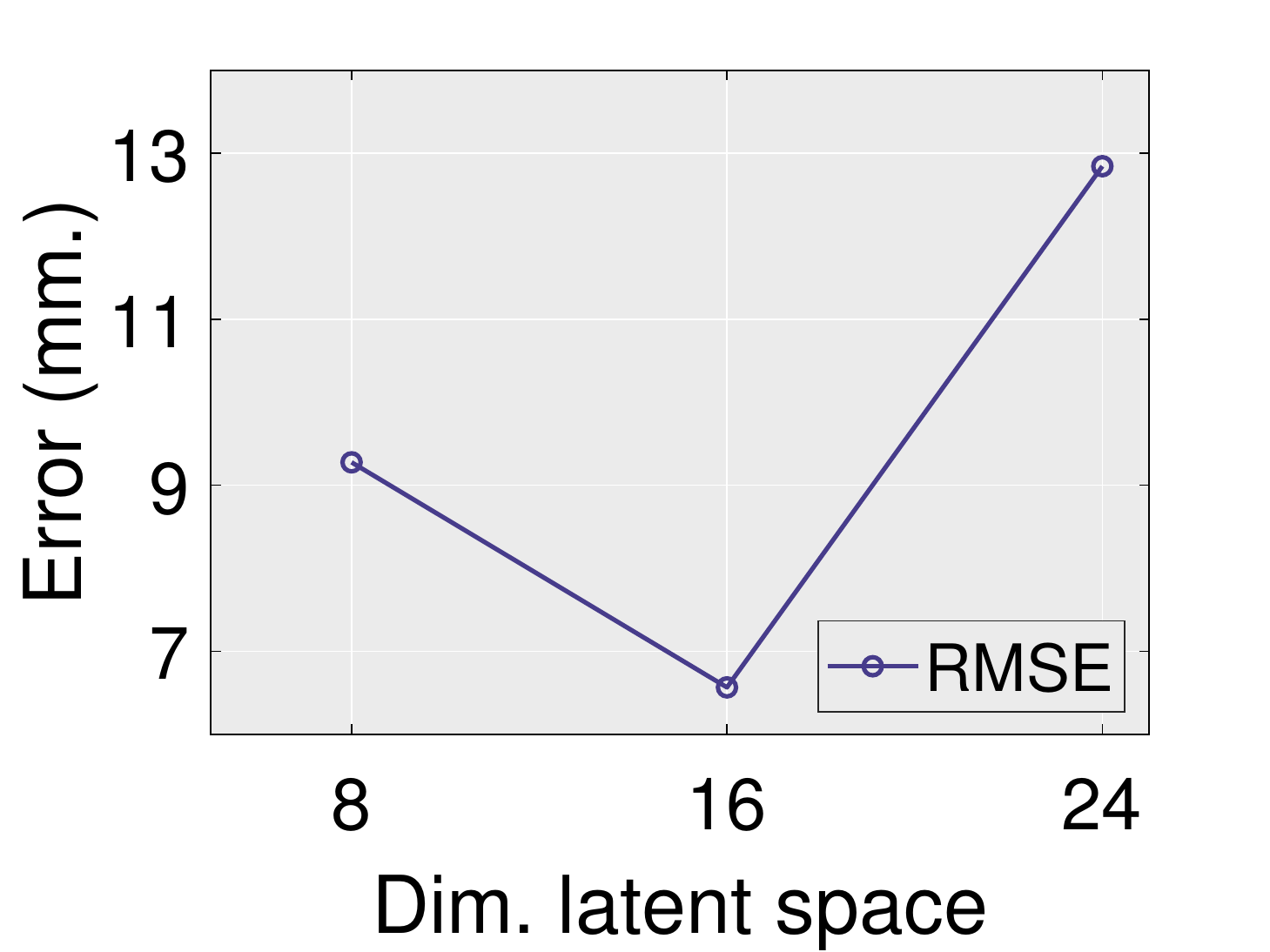} 
	}
	\caption{\small Evaluation of hyperparameters in the Transformer network\YL{, using the SKIRT dataset. }
		(Left) average error for the reconstructed results as a function of the number of input frames.
		(Right) error for the synthesized results under the condition of various dimensions of the latent layer.
	}
	\label{fig:hyperpara}
\end{figure}
\textbf{Network architecture.}
As shown in Fig.~\ref{fig:Transformer}, our transduction network consists of two components, namely convolutional \YL{encoders} to map coarse and fine mesh sequences into latent spaces for improved generalization capability, and the Transformer network for \YL{spatio-temporally} coherent deformation transduction.
The feature encoder module takes the 9-dimensional TS-ACAP features defined on vertices as input, followed by two convolutional layers with $tanh$ as the activation function. 
In the last convolutional layer we abandon the activation function, similar to \cite{tan2017autoencoder}.
A fully connected layer is used to map the output of the convolutional layers into a 16-dimensional latent space.
We train one encoder for coarse \YL{meshes} and another for fine \YL{meshes} separately.
For the DeformTransformer network, its input includes the embedded latent vectors from both \YL{the} coarse and fine domains.
The DeformTransformer network consists of sequential encoders and decoders, 
each \YL{including} a stack of 2 identical blocks with 8-head attention.
Different from variable length sequences used in natural language processing, we \YL{fix} the number of input frames \YL{(to 3 in our experiments)} since a motion sequence may include a thousand frames.
\YL{We perform experiments to evaluate the performance of our method with different settings.}
As shown in Fig.~\ref{fig:hyperpara} \YL{(left)}, using 3 input frames is found to perform well in our experiments.
We also evaluate the results generated with various dimensions of latent space shown in Fig. \ref{fig:hyperpara} \YL{(right)}.
When the dimension of latent space is larger than 16, the network can \YL{easily overfit}.
Thus we set the dimension of the latent space %
to 16, which is sufficient for all the examples in the paper.
\begin{table}[tb]
	\renewcommand\arraystretch{1.5}
	\caption{Quantitative comparison of reconstruction errors for unseen \YL{cloth animations} in several datasets. We compare our results with Chen {\itshape et al.} \cite{chen2018synthesizing} and Zurdo {\itshape et al.} \cite{zurdo2013wrinkles} with LR meshes as a reference. \YL{Three metrics, namely RMSE (Root Mean Squared Error), Hausdorff distance and STED (Spatio-Temporal Edge Difference)~\cite{Vasa2011perception} are used. Since LR meshes have different number of vertices from the ground truth HR mesh, we only calculate its Hausdorff distance.}}
 	\label{table:compare_zurdo_chen2}
	\centering 
	\begin{tabular}{ccccc} 
		\toprule[1.2pt]
		\multirow{3}{*}{Dataset} & \multirow{3}{*}{Methods} & \multicolumn{3}{c}{Metrics} \\ \cline{3-5}
		&                          & RMSE    & Hausdorff & STED  \\ 
		&                          & $\times 10^\YL{-2}$ $\downarrow$ & $\times 10^\YL{-2}$ $\downarrow$ & $\downarrow$ \\
		\hline \hline
		\multirow{4}{*}{TSHIRT}   &  LR    & -          & 0.59     & -     \\ \cline{2-5}
		&  Chen {\itshape et al.}  & 0.76   & 0.506    &  0.277     \\ \cline{2-5}
		&  Zurdo {\itshape et al.} & 1.04   & 0.480    &  0.281     \\ \cline{2-5}
		&  Our                     & \textbf{0.546}    &  \textbf{0.416}   & \textbf{0.0776}      \\ \hline  \hline
		\multirow{4}{*}{PANTS}   &  LR                   & -      &  0.761     & -     \\ \cline{2-5}
		&  Chen {\itshape et al.}  & 1.82              & 1.09 & 0.176   \\ \cline{2-5}
		&  Zurdo {\itshape et al.} & 1.89                 & 0.983& 0.151    \\ \cline{2-5}
		&  Our                    & \textbf{0.663} & \textbf{0.414}    & \textbf{0.0420}      \\ \hline \hline
		\multirow{4}{*}{SKIRT} &  LR                      & -      & 2.09     & -     \\ \cline{2-5}
		&  Chen {\itshape et al.}  & 1.93 &  1.31    & 0.562      \\ \cline{2-5}
		&  Zurdo {\itshape et al.} & 2.19 & 1.52     & 0.178      \\ \cline{2-5}
		&  Our                    & \textbf{0.685} & \textbf{0.681}    &  \textbf{0.0241}      \\ \hline  \hline
		\multirow{4}{*}{SHEET} 
		&  LR                      & -       & 2.61 &   -   \\ \cline{2-5}
		&  Chen {\itshape et al.}  & 4.37  & 2.60 &   0.155 \\ \cline{2-5}
		&  Zurdo {\itshape et al.} & 3.02  & 2.34 &   0.0672    \\ \cline{2-5}
		&  Our                     & \textbf{0.585} & \textbf{0.417}   &  \textbf{0.0262}     \\ \hline  \hline
		\multirow{4}{*}{DISK} &  LR & -      & 3.12    & -     \\ \cline{2-5}
		&  Chen {\itshape et al.}   & 7.03   & 2.27    &  0.244     \\ \cline{2-5}
		&  Zurdo {\itshape et al.}  & 11.40  &  2.23           & 0.502           \\ \cline{2-5}
		&  Our                      & \textbf{2.16}   & \textbf{1.30}   &  \textbf{0.0557 }     \\ 
		\bottomrule[1.2pt]
	\end{tabular}
\end{table}

\section{Results}\label{sec:results}
\subsection{Runtime Performance}
We implement our method on a \YL{computer with a} 2.50GHz \YL{4-Core} Intel CPU for coarse simulation and TS-ACAP extraction,
and \YL{an} NVIDIA GeForce\textsuperscript{\textregistered}~GTX 1080Ti GPU for fine TS-ACAP generation by the network and mesh coordinate reconstruction.
Table~\ref{table:runtime} shows average per-frame execution time of our method for various cloth datasets.
The execution time contains four parts: coarse simulation, TS-ACAP extraction, high-resolution TS-ACAP synthesis, and collision refinement. 
For reference, we also \YL{measure} the time of a CPU-based implementation of  high-resolution PBS using ARCSim \cite{Narain2012AAR}.
Our algorithm is $10\sim35$ times faster than the \YL{PBS} HR simulation.
The low computational cost of our method makes it suitable for the interactive applications. 

\begin{figure}[tb]
	\centering
	\setlength{\fboxrule}{0.5pt}
    \setlength{\fboxsep}{-0.01cm}
	\setlength{\tabcolsep}{0.00cm}  
    \renewcommand\arraystretch{0.01} 
 	\begin{tabular}{>{\centering\arraybackslash}m{0.2\linewidth}>{\centering\arraybackslash}m{0.2\linewidth}>{\centering\arraybackslash}m{0.2\linewidth}>{\centering\arraybackslash}m{0.2\linewidth}>{\centering\arraybackslash}m{0.2\linewidth}} 
     \includegraphics[width=\linewidth,  trim=17 0 37 0,clip]{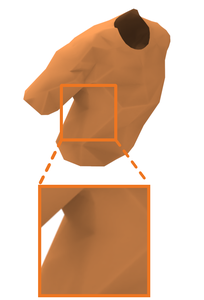} &  
     \includegraphics[width=\linewidth,  trim=17 0 37 0,clip]{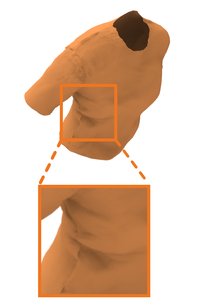} &  
     \includegraphics[width=\linewidth,  trim=17 0 37 0,clip]{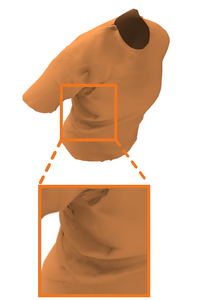} &  
     \includegraphics[width=\linewidth,  trim=17 0 37 0,clip]{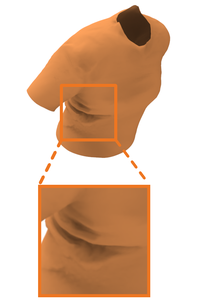} &  
     \includegraphics[width=\linewidth,  trim=17 0 37 0,clip]{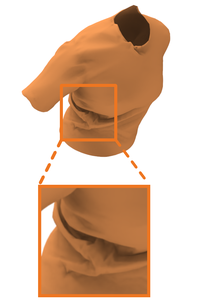} \\
     \includegraphics[width=\linewidth,  trim=17 0 37 0,clip]{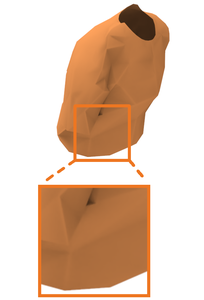} &  
     \includegraphics[width=\linewidth,  trim=17 0 37 0,clip]{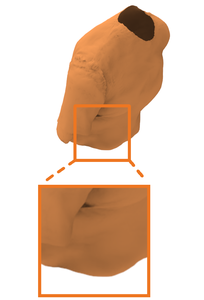} &  
     \includegraphics[width=\linewidth,  trim=17 0 37 0,clip]{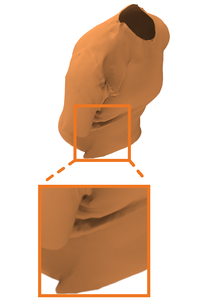} &  
     \includegraphics[width=\linewidth,  trim=17 0 37 0,clip]{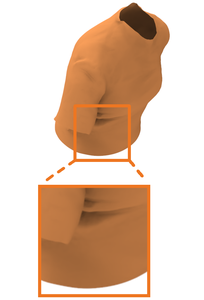} &  
     \includegraphics[width=\linewidth,  trim=17 0 37 0,clip]{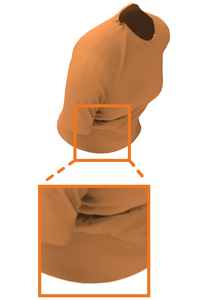} \\
     \includegraphics[width=\linewidth,  trim=17 0 37 0,clip]{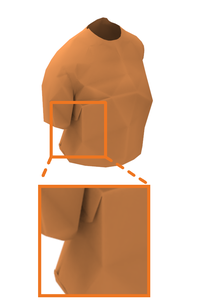} &  
     \includegraphics[width=\linewidth,  trim=17 0 37 0,clip]{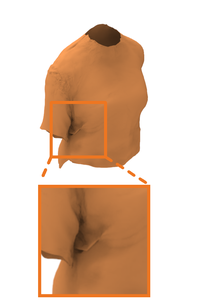} &  
     \includegraphics[width=\linewidth,  trim=17 0 37 0,clip]{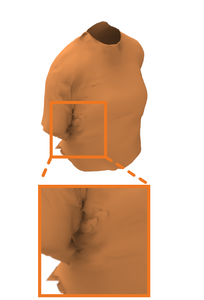} &  
     \includegraphics[width=\linewidth,  trim=17 0 37 0,clip]{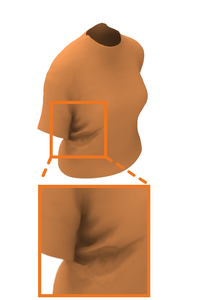} &  
     \includegraphics[width=\linewidth,  trim=17 0 37 0,clip]{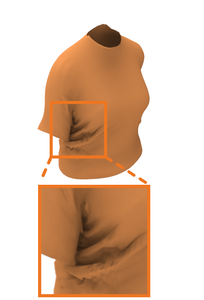} \\
     \includegraphics[width=\linewidth,  trim=17 0 37 0,clip]{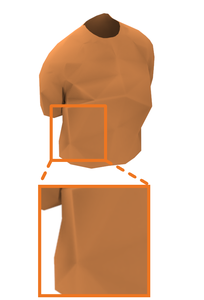} &  
     \includegraphics[width=\linewidth,  trim=17 0 37 0,clip]{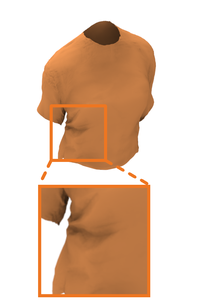} &  
     \includegraphics[width=\linewidth,  trim=17 0 37 0,clip]{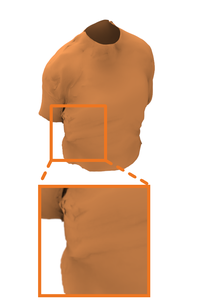} &  
     \includegraphics[width=\linewidth,  trim=17 0 37 0,clip]{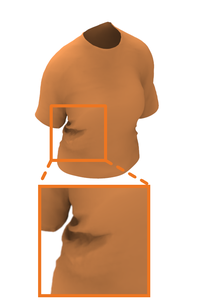} &  
     \includegraphics[width=\linewidth,  trim=17 0 37 0,clip]{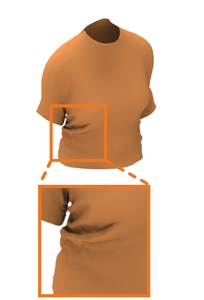} \\ 
      \vspace{0.3cm} \footnotesize (a) Input & \vspace{0.3cm} \hspace{-0.3cm} \footnotesize (b) Chen {\itshape et al.} & \vspace{0.3cm} \hspace{-0.2cm} \footnotesize (c) Zurdo {\itshape et al.} & \vspace{0.3cm} \footnotesize (d) Ours & \vspace{0.3cm} \footnotesize (e) GT  
	\end{tabular}
	\caption{Comparison of the reconstruction results for unseen data \YL{on the TSHIRT} dataset.
		(a) coarse simulation,
		(b) results of \cite{chen2018synthesizing},
		(c) results of \cite{zurdo2013wrinkles},
		(d) our results,
		(e) ground truth generated by PBS.
		Our method produces the detailed shapes of higher quality than Chen {\itshape et al.} and Zurdo {\itshape et al.}, see the folds and wrinkles in the close-ups. Chen {\itshape et al.} results suffer from seam line problems. The results of Zurdo {\itshape et al.} exhibit clearly noticeable artifacts.}
	\label{fig:comparetoothers_tshirt}
\end{figure}
 \begin{figure}[!htb]
	\centering
	\setlength{\fboxrule}{0.5pt}
    \setlength{\fboxsep}{-0.01cm}
	\setlength{\tabcolsep}{0.00cm}  
    \renewcommand\arraystretch{0.01} 
 	\begin{tabular}{>{\centering\arraybackslash}m{0.2\linewidth}>{\centering\arraybackslash}m{0.2\linewidth}>{\centering\arraybackslash}m{0.2\linewidth}>{\centering\arraybackslash}m{0.2\linewidth}>{\centering\arraybackslash}m{0.2\linewidth}} 
     \includegraphics[width=\linewidth, trim=28 0 28 5,clip]{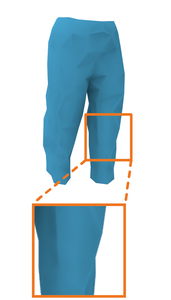} & 
 	\includegraphics[width=\linewidth, trim=28 0 28 5,clip]{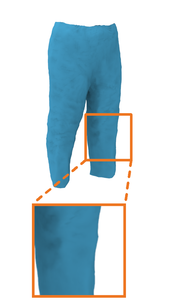} & 
 	\includegraphics[width=\linewidth, trim=28 0 28 5,clip]{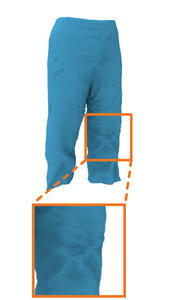} & 
 	\includegraphics[width=\linewidth, trim=28 0 28 5,clip]{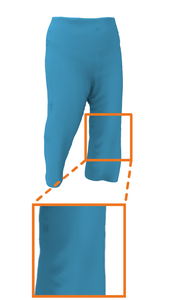} & 
 	\includegraphics[width=\linewidth, trim=28 0 28 5,clip]{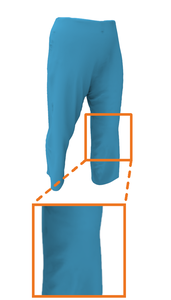} \\  
 	\includegraphics[width=\linewidth, trim=28 0 28 5,clip]{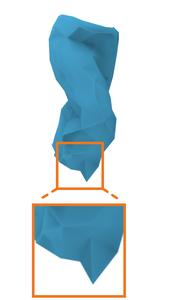} & 
 	\includegraphics[width=\linewidth, trim=28 0 28 5,clip]{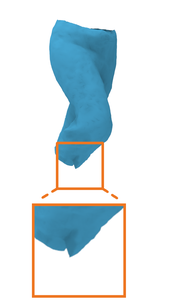} & 
 	\includegraphics[width=\linewidth, trim=28 0 28 5,clip]{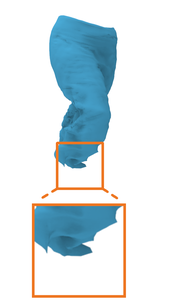} & 
 	\includegraphics[width=\linewidth, trim=28 0 28 5,clip]{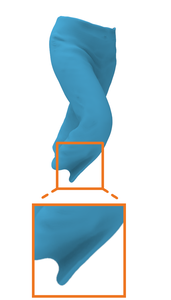} & 
 	\includegraphics[width=\linewidth, trim=28 0 28 5,clip]{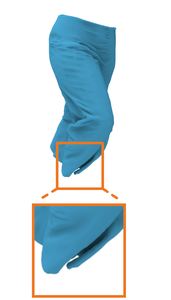} \\  
 	\includegraphics[width=\linewidth, trim=28 0 28 5,clip]{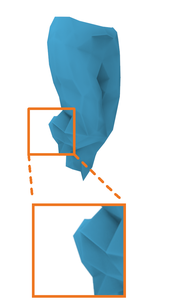} & 
 	\includegraphics[width=\linewidth, trim=28 0 28 5,clip]{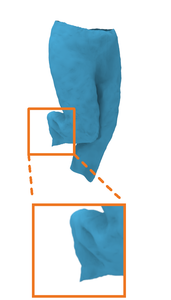} & 
 	\includegraphics[width=\linewidth, trim=28 0 28 5,clip]{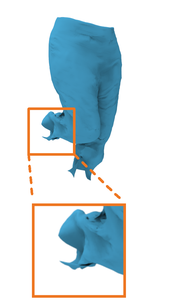} & 
 	\includegraphics[width=\linewidth, trim=28 0 28 5,clip]{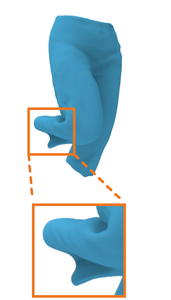} & 
 	\includegraphics[width=\linewidth, trim=28 0 28 5,clip]{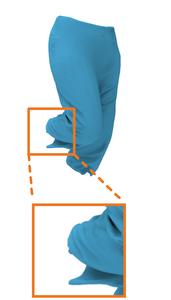} \\  
 	\includegraphics[width=\linewidth, trim=28 0 28 5,clip]{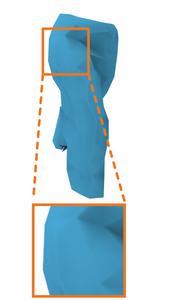} & 
 	\includegraphics[width=\linewidth, trim=28 0 28 5,clip]{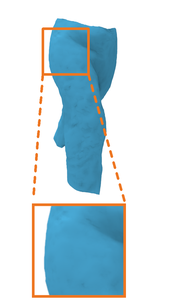} & 
 	\includegraphics[width=\linewidth, trim=28 0 28 5,clip]{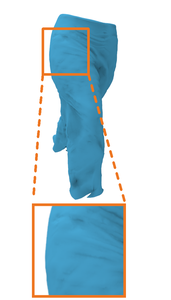} & 
 	\includegraphics[width=\linewidth, trim=28 0 28 5,clip]{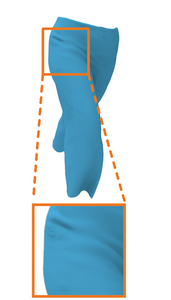} & 
 	\includegraphics[width=\linewidth, trim=28 0 28 5,clip]{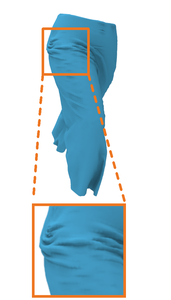} \\  
	 \vspace{0.3cm} \footnotesize (a) Input & \vspace{0.3cm} \hspace{-0.3cm} \footnotesize (b) Chen {\itshape et al.} & \vspace{0.3cm} \hspace{-0.2cm} \footnotesize (c) Zurdo {\itshape et al.} & \vspace{0.3cm} \footnotesize (d) Ours & \vspace{0.3cm} \footnotesize (e) GT  
	\end{tabular} 
	\caption{Comparison of the reconstruction results for unseen data in the PANTS dataset.
		(a) coarse simulation results,
		(b) results of \cite{chen2018synthesizing}, mainly smooth the coarse meshes and barely exhibit any wrinkles.
		(c) results of \cite{zurdo2013wrinkles}, have clear artifacts on examples where LR and HR meshes are not aligned well, \eg the trouser legs.
		(d) our results, ensures physically-reliable results.
		(e) ground truth generated by PBS.
	}
	\label{fig:comparetoothers_pants}
\end{figure}  
\begin{figure*}[htb]
	\centering
	\subfloat[Input]{ 
		\begin{minipage}[b]{0.11\linewidth} 
			\includegraphics[width=1.000000\linewidth, trim=45 45 45 45,clip]{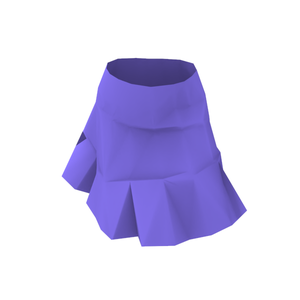} 
			\includegraphics[width=1.000000\linewidth, trim=45 45 45 45,clip]{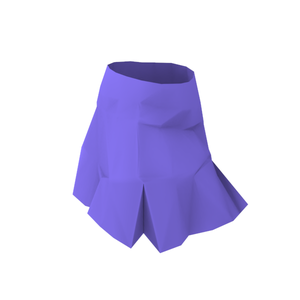} 
			\includegraphics[width=1.000000\linewidth, trim=45 45 45 45,clip]{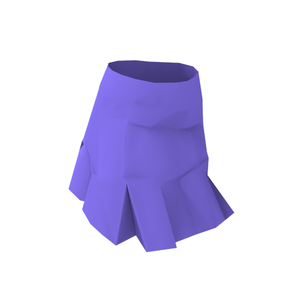} 
			\includegraphics[width=1.000000\linewidth, trim=45 45 45 45,clip]{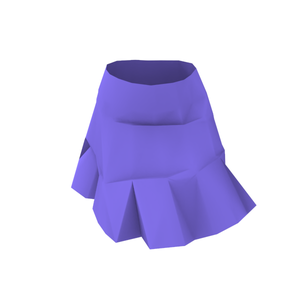} 
	\end{minipage}} 
	\subfloat[Chen {\itshape et al.}]{ 
		\begin{minipage}[b]{0.11\linewidth} 
			\includegraphics[width=1.000000\linewidth, trim=45 45 45 45,clip]{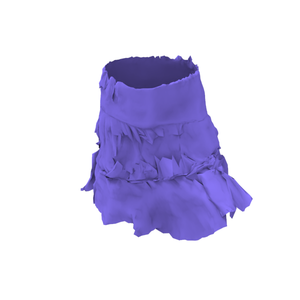} 
			\includegraphics[width=1.000000\linewidth, trim=45 45 45 45,clip]{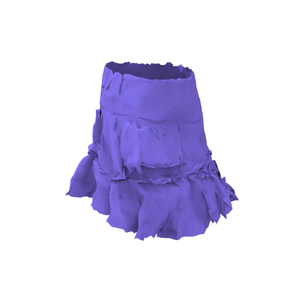} 
			\includegraphics[width=1.000000\linewidth, trim=45 45 45 45,clip]{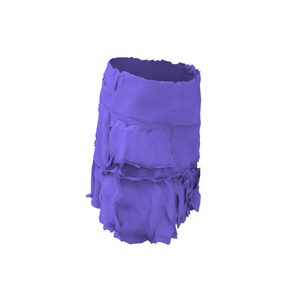} 
			\includegraphics[width=1.000000\linewidth, trim=45 45 45 45,clip]{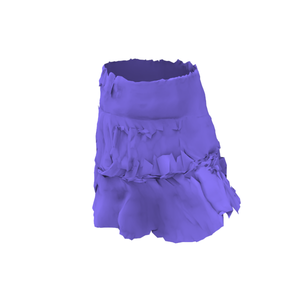} 
	\end{minipage}} 
		\begin{minipage}[b]{0.11\linewidth} 
			\includegraphics[width=1.000000\linewidth, trim=45 45 45 45 ,clip]{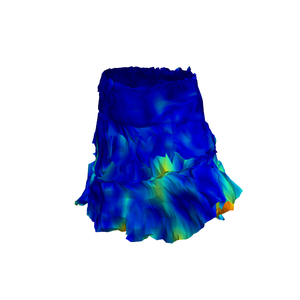} 
			\includegraphics[width=1.000000\linewidth, trim=45 45 45 45 ,clip]{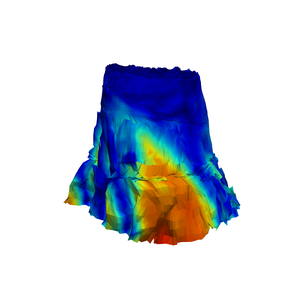} 
			\includegraphics[width=1.000000\linewidth, trim=45 45 45 45 ,clip]{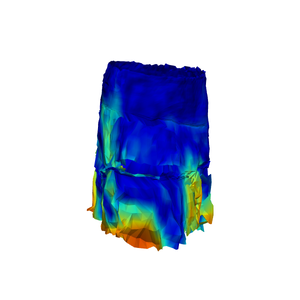} 
			\includegraphics[width=1.000000\linewidth, trim=45 45 45 45 ,clip]{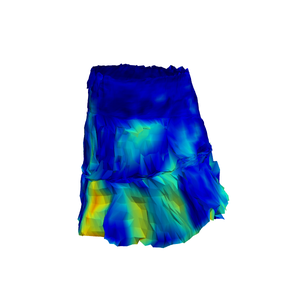} 
	\end{minipage}
	\subfloat[Zurdo {\itshape et al.}]{ 
		\begin{minipage}[b]{0.11\linewidth} 
			\includegraphics[width=1.000000\linewidth, trim=45 45 45 45,clip]{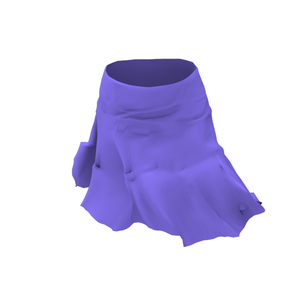} 
			\includegraphics[width=1.000000\linewidth, trim=45 45 45 45,clip]{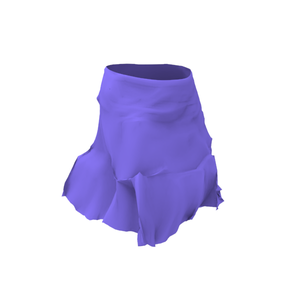} 
			\includegraphics[width=1.000000\linewidth, trim=45 45 45 45,clip]{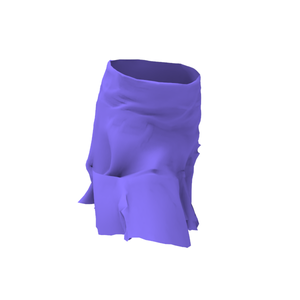} 
			\includegraphics[width=1.000000\linewidth, trim=45 45 45 45,clip]{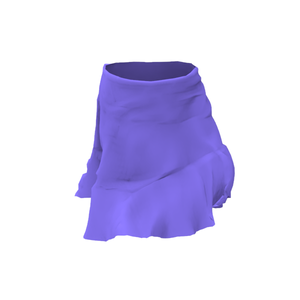} 
	\end{minipage}} 
		\begin{minipage}[b]{0.11\linewidth} 
			\includegraphics[width=1.000000\linewidth, trim=45 45 45 45,clip]{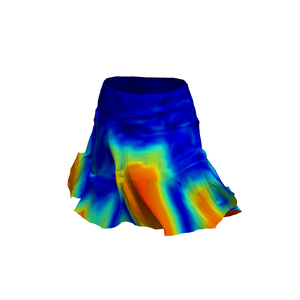} 
			\includegraphics[width=1.000000\linewidth, trim=45 45 45 45,clip]{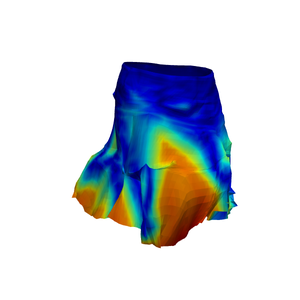} 
			\includegraphics[width=1.000000\linewidth, trim=45 45 45 45,clip]{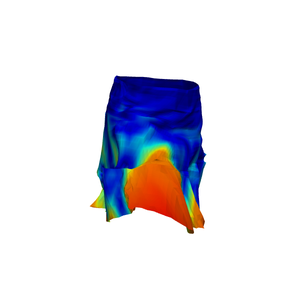} 
			\includegraphics[width=1.000000\linewidth, trim=45 45 45 45,clip]{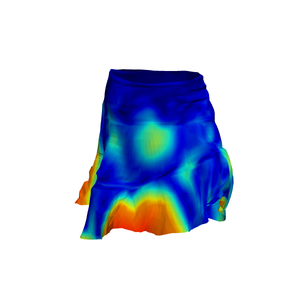} 
	\end{minipage}
	\subfloat[Ours]{ 
		\begin{minipage}[b]{0.11\linewidth} 
			\includegraphics[width=1.000000\linewidth, trim=45 45 45 45,clip]{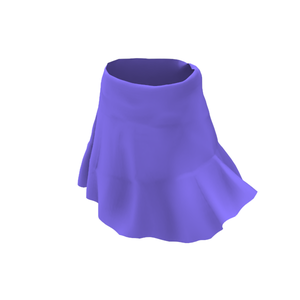} 
			\includegraphics[width=1.000000\linewidth, trim=45 45 45 45,clip]{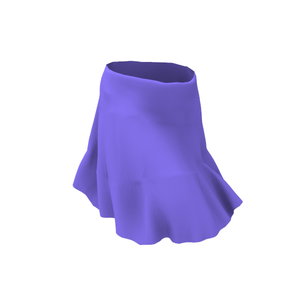} 
			\includegraphics[width=1.000000\linewidth, trim=45 45 45 45,clip]{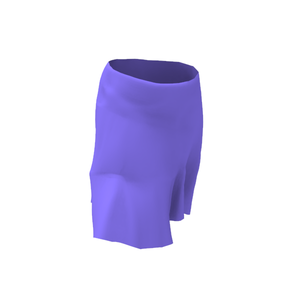} 
			\includegraphics[width=1.000000\linewidth, trim=45 45 45 45,clip]{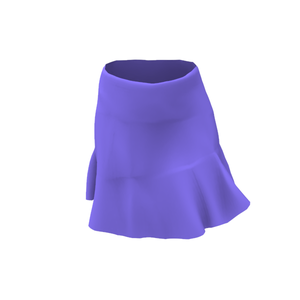} 
	\end{minipage}}
		\begin{minipage}[b]{0.11\linewidth} 
			\includegraphics[width=1.000000\linewidth, trim=45 45 45 45,clip]{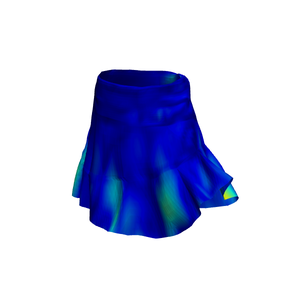} 
			\includegraphics[width=1.000000\linewidth, trim=45 45 45 45,clip]{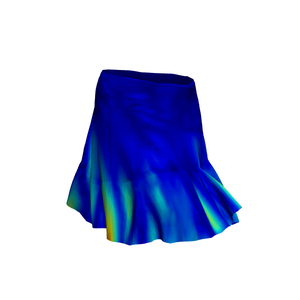} 
			\includegraphics[width=1.000000\linewidth, trim=45 45 45 45,clip]{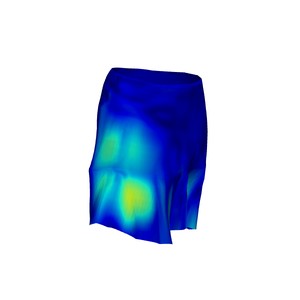} 
			\includegraphics[width=1.000000\linewidth, trim=45 45 45 45,clip]{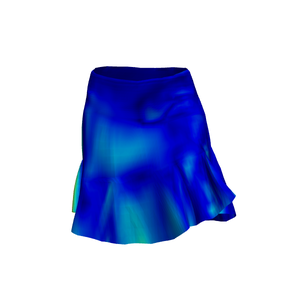} 
	\end{minipage}  
	\subfloat[GT]{ 
		\begin{minipage}[b]{0.11\linewidth} 
			\includegraphics[width=1.000000\linewidth, trim=45 45 45 45,clip]{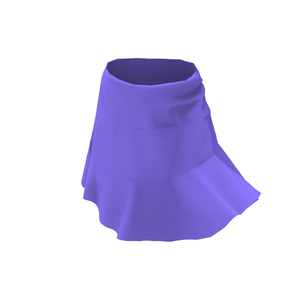} 
			\includegraphics[width=1.000000\linewidth, trim=45 45 45 45,clip]{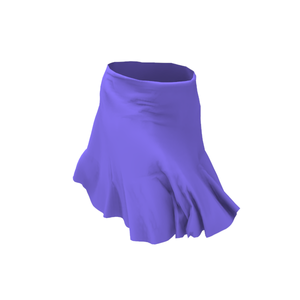} 
			\includegraphics[width=1.000000\linewidth, trim=45 45 45 45,clip]{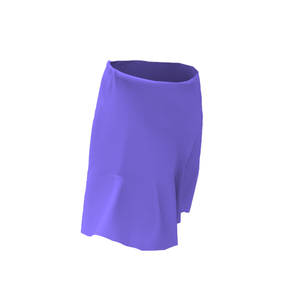} 
			\includegraphics[width=1.000000\linewidth, trim=45 45 45 45,clip]{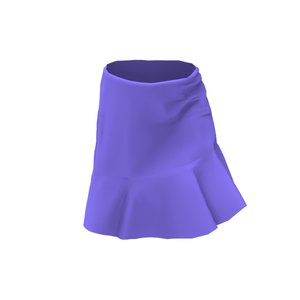} 
	\end{minipage}}
	\begin{minipage}[b]{0.08\linewidth} 
		\includegraphics[width=1.000000\linewidth, trim=0 0 0 0,clip]{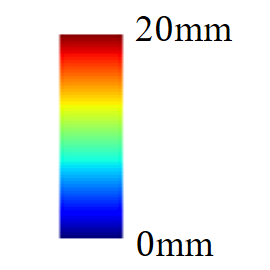}
	\end{minipage}
	
	\caption{Comparison of the reconstruction results for unseen data in the SKIRT dataset.
		(a) the coarse simulation,
		(b) the results of \cite{chen2018synthesizing},
		(c) the results of \cite{zurdo2013wrinkles},
		(d) our results,
		(e) the ground truth generated by PBS.
	The reconstruction accuracy is qualitatively showed as a difference map. 
	Reconstruction errors are color-coded and warmer colors indicate larger errors. Our method leads to significantly lower reconstruction errors. }
	\label{fig:comparetoothers_skirt}
\end{figure*}

\subsection{\YL{Fine Detail} Synthesis Results and Comparisons}
We now demonstrate our method using various \YL{detail enhancement}
examples \YL{both} quantitatively and qualitatively, \YL{including added wrinkles and rich dynamics.}
Using detailed meshes generated by PBS as ground truth, we compare our results with physics-based coarse simulations, our implementation of a deep learning-based method \cite{chen2018synthesizing} and a conventional machine learning-based method \cite{zurdo2013wrinkles}.

For quantitative comparison, we use \YL{three} metrics: Root Mean Squared Error (RMSE), Hausdorff distance as well as spatio-temporal edge difference (STED) \cite{Vasa2011perception} designed for motion sequences with a focus on `perceptual’ error of models.
The results are shown in Table~\ref{table:compare_zurdo_chen2}.
Note that \YL{for the datasets from the top to bottom in the table,} the Hausdorff \YL{distances} between LR meshes and the ground truth are increasing. \YL{This} tendency is in accordance with the deformation range from tighter T-shirts and pants to skirts and square/disk tablecloth with higher degrees \YL{of freedom}.
Since the vertex position representation cannot handle rotations well, the larger scale the models deform, the more artifacts Chen {\itshape et al.} \cite{chen2018synthesizing} and Zurdo {\itshape et al.} \cite{zurdo2013wrinkles} would \YL{bring in} in the reconstructed models, \YL{leading  to increased} RMSE and Hausdorff distances.  
The results indicate that our method has better reconstruction results \YL{quantitatively} than the compared methods \YL{on} the 5 datasets with \YL{all the three} metrics.
Especially \YL{for} the SKIRT, SHEET and DISK \YL{datasets} which \YL{contain} loose cloth \YL{and hence larger and richer deformation}, our \YL{method} outperforms \YL{existing methods significantly} since tracking between coarse and fine meshes \YL{is} not required in our algorithm.

\begin{figure}[tb]
	\centering
	\setlength{\fboxrule}{0.5pt}
    \setlength{\fboxsep}{-0.01cm}
	\setlength{\tabcolsep}{0.00cm}  
    \renewcommand\arraystretch{0.01} 
 	\begin{tabular}{>{\centering\arraybackslash}m{0.2\linewidth}>{\centering\arraybackslash}m{0.2\linewidth}>{\centering\arraybackslash}m{0.2\linewidth}>{\centering\arraybackslash}m{0.2\linewidth}>{\centering\arraybackslash}m{0.2\linewidth}} 
     \includegraphics[width=\linewidth, trim=25 0 50 6,clip]{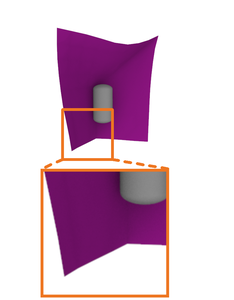} & 
 	\includegraphics[width=\linewidth, trim=25 0 50 6,clip]{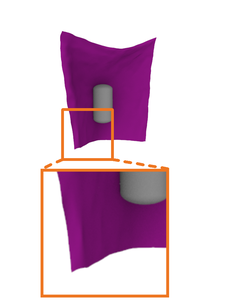} & 
 	\includegraphics[width=\linewidth, trim=25 0 50 6,clip]{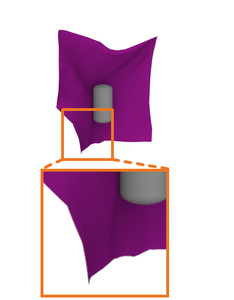} & 
 	\includegraphics[width=\linewidth, trim=25 0 50 6,clip]{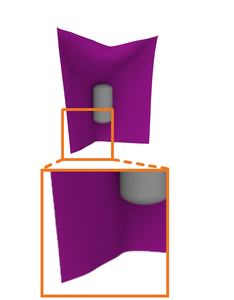} & 
 	\includegraphics[width=\linewidth, trim=25 0 50 6,clip]{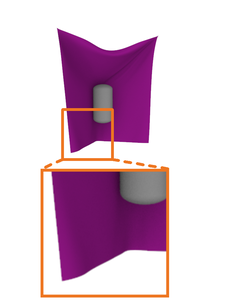}\\ 
 	\includegraphics[width=\linewidth, trim=25 0 50 6,clip]{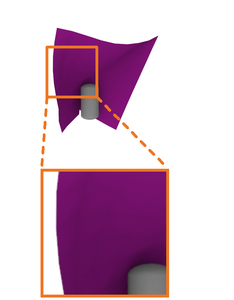} & 
 	\includegraphics[width=\linewidth, trim=25 0 50 6,clip]{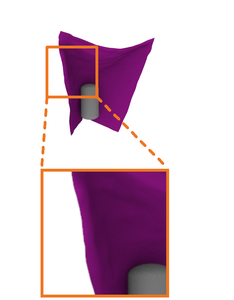} & 
 	\includegraphics[width=\linewidth, trim=25 0 50 6,clip]{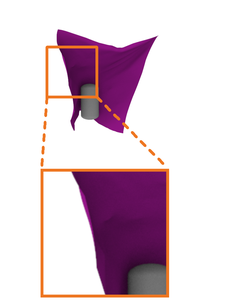} & 
 	\includegraphics[width=\linewidth, trim=25 0 50 6,clip]{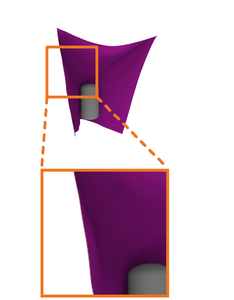} & 
 	\includegraphics[width=\linewidth, trim=25 0 50 6,clip]{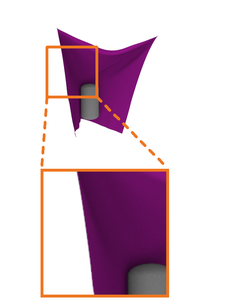} \\ 
 	\includegraphics[width=\linewidth, trim=25 0 50 6,clip]{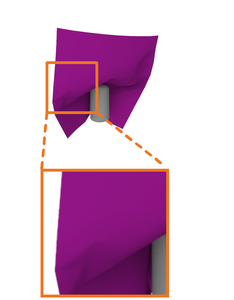} & 
 	\includegraphics[width=\linewidth, trim=25 0 50 6,clip]{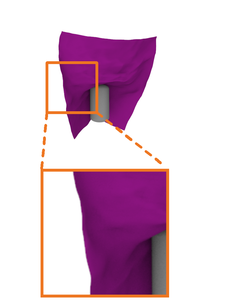} & 
 	\includegraphics[width=\linewidth, trim=25 0 50 6,clip]{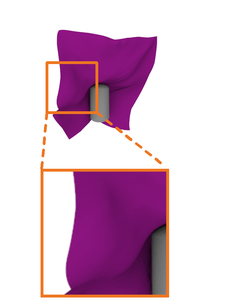} & 
 	\includegraphics[width=\linewidth, trim=25 0 50 6,clip]{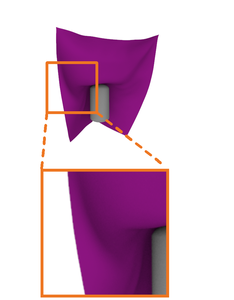} & 
 	\includegraphics[width=\linewidth, trim=25 0 50 6,clip]{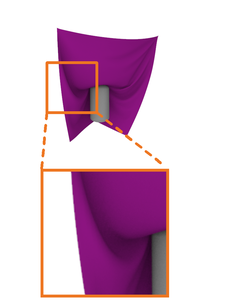} \\ 
 	\includegraphics[width=\linewidth, trim=25 0 50 6,clip]{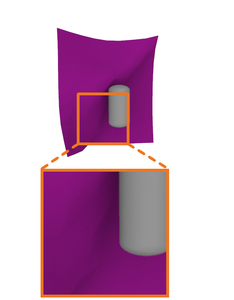} & 
 	\includegraphics[width=\linewidth, trim=25 0 50 6,clip]{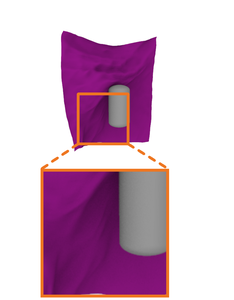} & 
 	\includegraphics[width=\linewidth, trim=25 0 50 6,clip]{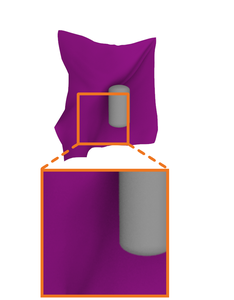} & 
 	\includegraphics[width=\linewidth, trim=25 0 50 6,clip]{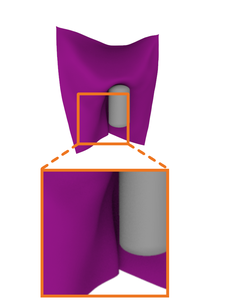} & 
 	\includegraphics[width=\linewidth, trim=25 0 50 6,clip]{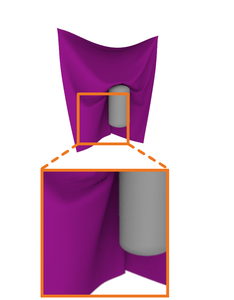} \\ 
	 \vspace{0.3cm} \footnotesize (a) Input & \vspace{0.3cm} \hspace{-0.3cm} \footnotesize (b) Chen {\itshape et al.} & \vspace{0.3cm} \hspace{-0.2cm} \footnotesize (c) Zurdo {\itshape et al.} & \vspace{0.3cm} \footnotesize (d) Ours & \vspace{0.3cm} \footnotesize (e) GT  
	\end{tabular}
	\caption{Comparison of the reconstruction results for unseen data in the SHEET dataset.
		(a) the coarse simulation,
		(b) the results of \cite{chen2018synthesizing}, with inaccurate and
rough wrinkles different from the GT.
		(c) the results of \cite{zurdo2013wrinkles}, show similar global shapes to coarse meshes with some wrinkles and unexpected sharp corner.
		(d) our results, show mid-scale wrinkles and similar global deformation as GT.
		(e) the ground truth generated by PBS.}
	\label{fig:comparetoothers_crashball}
	\vspace{-0.2cm}
\end{figure}   
\begin{figure}[tb]
	\centering
	\setlength{\fboxrule}{0.5pt}
    \setlength{\fboxsep}{-0.01cm}
	\setlength{\tabcolsep}{0.00cm}  
    \renewcommand\arraystretch{0.001} 
 	\begin{tabular}{>{\centering\arraybackslash}m{0.2\linewidth}>{\centering\arraybackslash}m{0.2\linewidth}>{\centering\arraybackslash}m{0.2\linewidth}>{\centering\arraybackslash}m{0.2\linewidth}>{\centering\arraybackslash}m{0.2\linewidth}} 
			 \includegraphics[width=\linewidth, trim=42 0 30 30,clip]{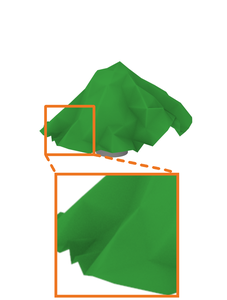} &
			 \includegraphics[width=\linewidth, trim=42 0 30 30,clip]{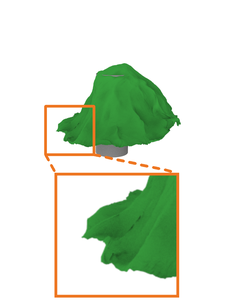} &
			 \includegraphics[width=\linewidth, trim=42 0 30 30,clip]{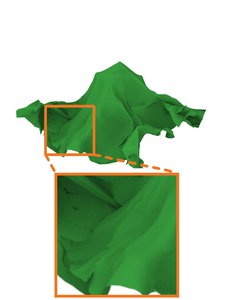} &
			 \includegraphics[width=\linewidth, trim=42 0 30 30,clip]{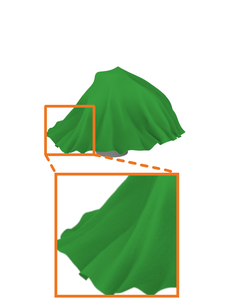} &
			 \includegraphics[width=\linewidth, trim=42 0 30 30,clip]{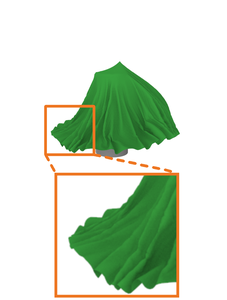} \\
			 \includegraphics[width=\linewidth, trim=54 0 18 30,clip]{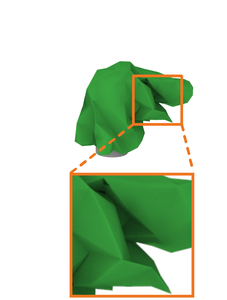} &
			 \includegraphics[width=\linewidth, trim=54 0 18 30,clip]{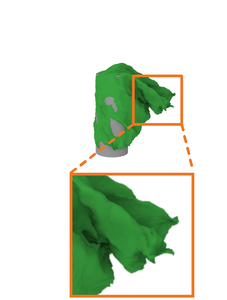} &
			 \includegraphics[width=\linewidth, trim=54 0 18 30,clip]{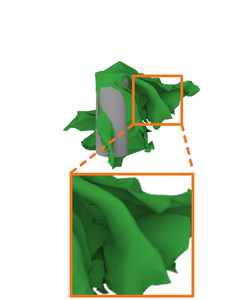} &
			 \includegraphics[width=\linewidth, trim=54 0 18 30,clip]{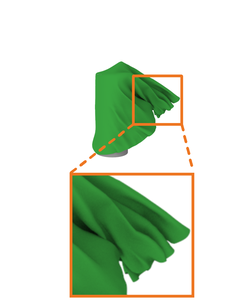} &
			 \includegraphics[width=\linewidth, trim=54 0 18 30,clip]{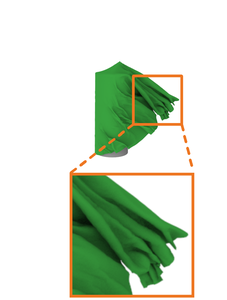} \\
			 \includegraphics[width=\linewidth, trim=54 0 18 30,clip]{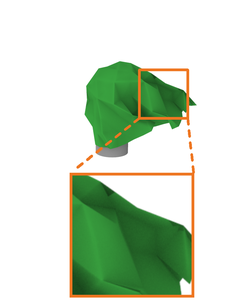} &
			 \includegraphics[width=\linewidth, trim=54 0 18 30,clip]{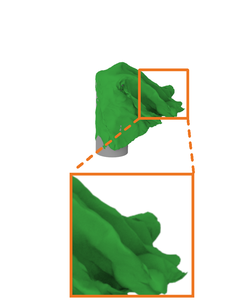} &
			 \includegraphics[width=\linewidth, trim=54 0 18 30,clip]{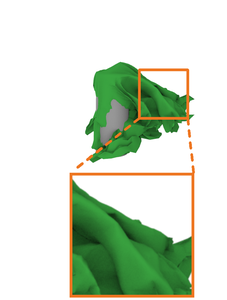} &
			 \includegraphics[width=\linewidth, trim=54 0 18 30,clip]{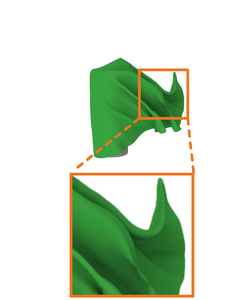} &
			 \includegraphics[width=\linewidth, trim=54 0 18 30,clip]{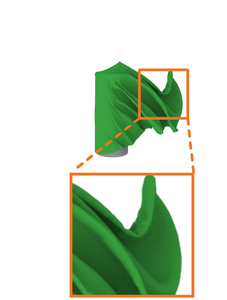} \\
			 \includegraphics[width=\linewidth, trim=54 0 18 30,clip]{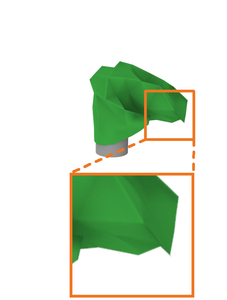} &
			 \includegraphics[width=\linewidth, trim=54 0 18 30,clip]{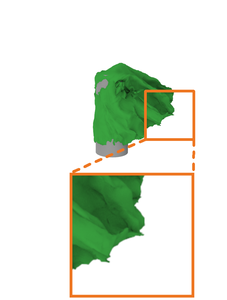} &
			 \includegraphics[width=\linewidth, trim=54 0 18 30,clip]{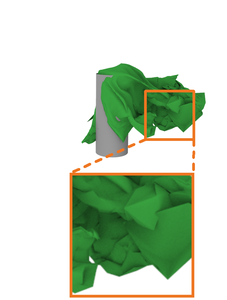} &
			 \includegraphics[width=\linewidth, trim=54 0 18 30,clip]{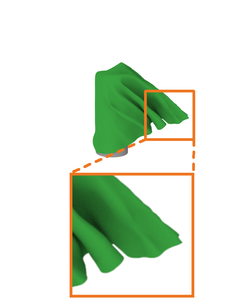} &
			 \includegraphics[width=\linewidth, trim=54 0 18 30,clip]{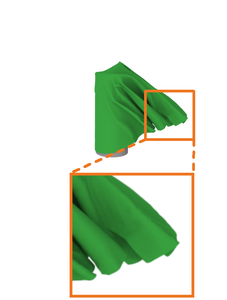} \\ 
			 \vspace{0.3cm} \footnotesize (a) Input & \vspace{0.3cm} \hspace{-0.3cm} \footnotesize (b) Chen {\itshape et al.} & \vspace{0.3cm} \hspace{-0.2cm} \footnotesize (c) Zurdo {\itshape et al.} & \vspace{0.3cm} \footnotesize (d) Ours & \vspace{0.3cm} \footnotesize (e) GT  
	\end{tabular}
	\caption{Comparison of the reconstruction results for unseen data in the DISK dataset.
		(a) the coarse simulation,
		(b) the results of \cite{chen2018synthesizing}, cannot reconstruct credible shapes. 
		(c) the results of \cite{zurdo2013wrinkles}, show apparent artifacts near the flying tails since no tracking constraints applied.
		(d) our results, reproduce large-scale deformations, see the tail of the disk flies like a fan in the wind.
		(e) the ground truth generated by PBS.}
	\label{fig:comparetoothers_disk}
\end{figure}    

\YL{We further make qualitative comparisons on the 5 datasets.}
Fig. \ref{fig:comparetoothers_tshirt} shows \YL{detail synthesis results} on the TSHIRT dataset.
The first and second 
rows 
are from \YL{sequence} 06\_08, a woman dribbling the basketball sideways and the \YL{last two rows} are from \YL{sequence} 08\_11, a walking woman.
In this dataset of tight t-shirts on human bodies, Chen {\itshape et al.} \cite{chen2018synthesizing}, Zurdo {\itshape et al.} \cite{zurdo2013wrinkles} and our method are able to reconstruct the garment model completely with mid-scale wrinkles.
However, Chen {\itshape et al.} \cite{chen2018synthesizing} suffer from the seam line problems due to \YL{the use of geometry image representation}. 
A geometry image is a parametric sampling of the shape, which is \YL{made a topological disk by cutting through some seams.} 
The boundary of the disk needs to be fused so that the reconstructed mesh has the original topology.
\YL{The super-resolved geometry image corresponding to high-resolution cloth animations are not entirely accurate, and as a result the fused boundaries no longer match exactly, }
\eg clear seam lines on the shoulder and crooked boundaries on the left side of the waist \YL{for the examples} in Fig.~\ref{fig:comparetoothers_tshirt} (b)),
\YL{while} our method \YL{produces} better results than \cite{chen2018synthesizing} and \cite{zurdo2013wrinkles} which have  \YL{artifacts of unsmooth surfaces}.

Fig. \ref{fig:comparetoothers_pants} shows comparative results of the animations of pants on a fixed body shape while changing the body pose over time. 
The results of \cite{chen2018synthesizing} \YL{mainly} smooth the coarse meshes and barely exhibit \YL{any} wrinkles.
Zurdo {\itshape et al.} \cite{zurdo2013wrinkles} utilize tracking algorithms to ensure the %
\YL{close alignment}
between coarse and fine meshes, and thus the fine meshes are constrained \YL{and do not exhibit the behavior of full physics-based simulation.}
\YL{So on the PANTS dataset,} the results of \cite{zurdo2013wrinkles} have clear artifacts on examples \YL{where} LR and HR meshes are not aligned well, \eg the trouser legs.
Different from the two compared methods \YL{that reconstruct displacements} or local coordinates, 
our method \YL{uses} deformation-based features in both encoding and decoding \YL{phases} which \YL{does not suffer from such restrictions and ensures physically-reliable results.}

For looser garments like \YL{skirts}, we show comparison results in Fig. \ref{fig:comparetoothers_skirt}, with color coding to highlight the differences between synthesized results and the ground truth.
Our method successfully reconstructs the swinging skirt \YL{caused by} the body motion (see the small wrinkles on the waist and the \YL{medium-level} folds on the skirt \YL{hem}).
Chen {\itshape et al.} are able to reconstruct the overall shape of the skirt, however there are many small unsmooth \YL{triangles leading to noisy shapes}
due to the 3D coordinate representation with untracked fine meshes with abundant wrinkles.
This leads to unstable animation, please see the accompanying video.
The results of \cite{zurdo2013wrinkles} have some problems of the global deformation, see the directions of the skirt hem and the large highlighted area in the color map.
Our learned \YL{detail} synthesis model provides better visual quality for shape generation \YL{and the generated results look} closer to the ground truth.
 
Instead of garments dressed on human bodies, we additionally show some results of free-flying tablecloth. 
The comparison of the testing results \YL{on} the SHEET dataset are shown in Fig.~\ref{fig:comparetoothers_crashball}.
The results of \cite{chen2018synthesizing} show inaccurate and rough wrinkles different from the ground truth. 
For hanging sheets in the results of \cite{zurdo2013wrinkles}, the global shapes are more like coarse \YL{meshes} with some wrinkles and unexpected sharp corners, \eg the left side in the last row of Fig. \ref{fig:comparetoothers_crashball} (c),
while ours show \YL{mid-scale} wrinkles and similar global deformation \YL{as} the high-resolution meshes. 

As for the DISK dataset, from the visual results in Fig.~\ref{fig:comparetoothers_disk}, we can see that Chen {\itshape et al.} \cite{chen2018synthesizing} and Zurdo {\itshape et al.} \cite{zurdo2013wrinkles} cannot handle large-scale rotations well and cannot reconstruct credible shapes in such cases. 
\gl{Especially for Zurdo {\itshape et al.} \cite{zurdo2013wrinkles}, the impact of tracking is significant for their algorithm.}
They can reconstruct the top
and part of tablecloth near the cylinder, but the flying tails have apparent artifacts.  
Our algorithm does not have such drawbacks.
Notice how our method successfully reproduces ground-truth deformations, including the overall drape (\ie, how the tail of the disk flies like a fan in the wind) and mid-scale wrinkles.

\begin{table}[!htb]
	\renewcommand\arraystretch{1.5}
	\caption{User study results on cloth \YL{detail} synthesis. We show the average ranking score of the three methods: Chen {\itshape et al.} \cite{chen2018synthesizing}, Zurdo {\itshape et al.} \cite{zurdo2013wrinkles}, and ours. The
		ranking ranges from 1 (the best) to 3 (the worst). The results are calculated
		based on 320 trials. We see that our method achieves the best in terms of
		wrinkles, temporal stability \YL{and overall quality}.}
	\label{table:userstudy}
	\centering 
	\begin{tabular}{cccc}
		\toprule[1.2pt]  
		Method                    & Wrinkles   & Temporal stability & Overall   \\ \hline 
		Chen {\itshape et al.}    & 2.184 & 2.1258 &2.1319\\ \hline  
		Zurdo {\itshape et al.}   & 2.3742 & 2.5215 & 2.4877\\ \hline         
		Ours                      & \textbf{1.4417} & \textbf{1.3528} & \textbf{1.3804}  \\
		\bottomrule[1.2pt]
	\end{tabular}
\end{table}
\gl{We further conduct a user study to evaluate the stability and realistic of the synthesized dense mesh dynamics. 32 volunteers are involved for this user study.}
For every question, we give one sequence and 5 images of coarse meshes as references, \YL{and} then let the user rank the corresponding outputs from  Chen {\itshape et al.} \cite{chen2018synthesizing}, Zurdo {\itshape et al.} \cite{zurdo2013wrinkles} and ours according to three different criteria (wrinkles, temporal stability and overall). 
We shuffle the order of the algorithms each time we exhibit the question and show shapes from the three methods randomly \YL{to avoid bias}. 
We show the results of the user study in Table \ref{table:userstudy}, where we observe that our generated \YL{shapes} perform the best on all three criteria. 

\begin{table}[tb]
	\renewcommand\arraystretch{1.5}
	\caption{Per-vertex error (RMSE) on synthesized shapes with different feature representations: 3D coordinates, ACAP and TS-ACAP.}
	\label{table:feature_compare}
	\centering
	\begin{tabular}{cccccc}
		\toprule[1.2pt]
		Dataset        & TSHIRT      & PANTS    &  	SKIRT   & SHEET & DISK      \\ \hline
		3D coordinates &  0.0101      & 0.0193   &    0.00941 &  0.00860   &  0.185 \\ \hline
		ACAP           &  0.00614     & 0.00785  &   0.00693  &  0.00606   &   0.0351     \\ \hline
		TS-ACAP          &  \textbf{0.00546}     & \textbf{0.00663}  &    \textbf{0.00685} &   \textbf{0.00585} & \textbf{0.0216}\\ 
		\bottomrule[1.2pt]
	\end{tabular}
\end{table}
\begin{figure}[tb]
	\centering
	\setlength{\fboxrule}{0.5pt}
    \setlength{\fboxsep}{-0.01cm}
	\setlength{\tabcolsep}{0.00cm}  
    \renewcommand\arraystretch{0.001}
 	\begin{tabular}{>{\centering\arraybackslash}m{0.25\linewidth}>{\centering\arraybackslash}m{0.25\linewidth}>{\centering\arraybackslash}m{0.25\linewidth}>{\centering\arraybackslash}m{0.25\linewidth}}
 	\includegraphics[width=1.000000\linewidth, trim=63 0 0 0,clip]{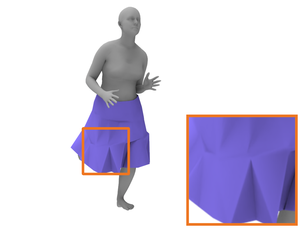} &
 	\includegraphics[width=1.000000\linewidth, trim=63 0 0 0,clip]{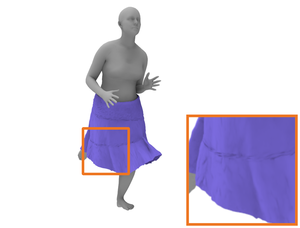} &
 	\includegraphics[width=1.000000\linewidth, trim=63 0 0 0,clip]{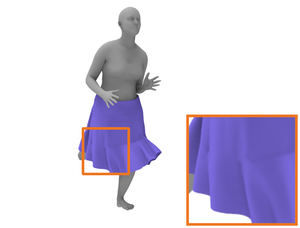} &
 	\includegraphics[width=1.000000\linewidth, trim=63 0 0 0,clip]{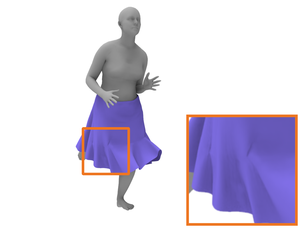} \\
 	\includegraphics[width=1.000000\linewidth, trim=63 0 0 0,clip]{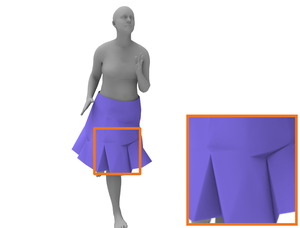} &
 	\includegraphics[width=1.000000\linewidth, trim=63 0 0 0,clip]{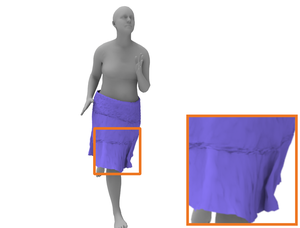} &
 	\includegraphics[width=1.000000\linewidth, trim=63 0 0 0,clip]{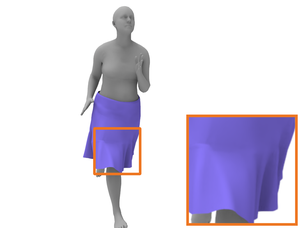} &
 	\includegraphics[width=1.000000\linewidth, trim=63 0 0 0,clip]{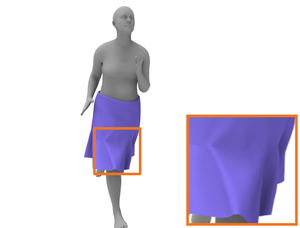} \\
 	\includegraphics[width=1.000000\linewidth, trim=63 0 0 0,clip]{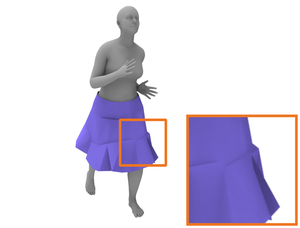} &
 	\includegraphics[width=1.000000\linewidth, trim=63 0 0 0,clip]{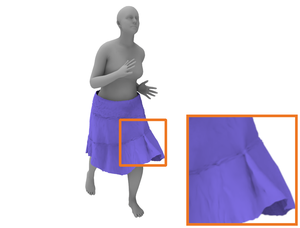} &
 	\includegraphics[width=1.000000\linewidth, trim=63 0 0 0,clip]{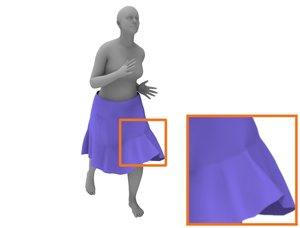} &
 	\includegraphics[width=1.000000\linewidth, trim=63 0 0 0,clip]{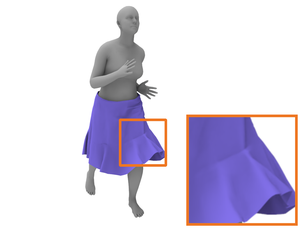} \\ 
 	\vspace{0.3cm} \small (a) Input & \vspace{0.3cm}\small (b) Coordinates & \vspace{0.3cm}\small (c) Ours & \vspace{0.3cm}\small (d) GT
  \end{tabular} 
	\caption{The evaluation of the TS-ACAP feature in our detail synthesis method. 
		(a) input coarse \YL{shapes},
		(b) the results using 3D coordinates, which can be clearly seen the rough appearance, unnatural deformation and some artifacts, especially in the highlighted regions with details shown in the close-ups.
		(c) our results, which show smooth looks and the details are more similar to the GT.
		(d)	ground truth.
		 }
	\label{fig:ablationstudy_coordiniates_skirt}
\end{figure}
\begin{figure}[htb]
	\centering
	\setlength{\tabcolsep}{0.05cm}  
    \renewcommand\arraystretch{0.001}
 	\begin{tabular}{>{\centering\arraybackslash}m{0.02\linewidth}>{\centering\arraybackslash}m{0.31\linewidth}>{\centering\arraybackslash}m{0.31\linewidth}>{\centering\arraybackslash}m{0.31\linewidth}}
 	 \rotatebox{90}{\small ACAP} &
 		\includegraphics[width=\linewidth, trim=90 0 0 60,clip]{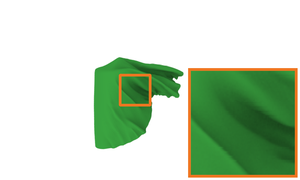} &
 		\includegraphics[width=\linewidth, trim=90 0 0 60,clip]{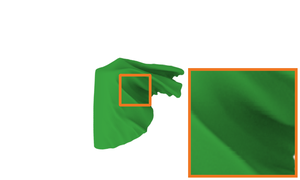} &
 		\includegraphics[width=\linewidth, trim=90 0 0 60,clip]{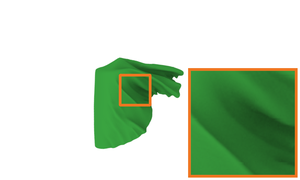} \\
 	\rotatebox{90}{\small TS-ACAP}  &
 		\includegraphics[width=\linewidth, trim=90 0 0 60,clip]{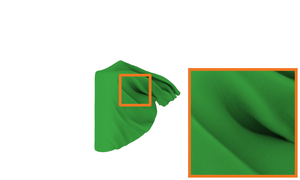} &
 		\includegraphics[width=\linewidth, trim=90 0 0 60,clip]{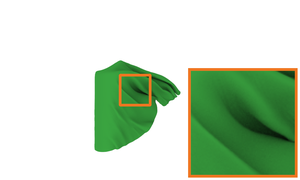} &
 		\includegraphics[width=\linewidth, trim=90 0 0 60,clip]{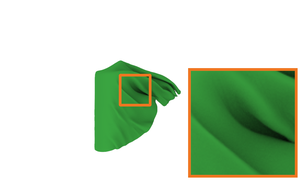} \\  
     \vspace{0.3cm}    & \vspace{0.3cm}  \small $t = 103$ & \vspace{0.3cm} \small $t = 104$  & \vspace{0.3cm} \small $t = 105$ 
	\end{tabular} 
	\caption{
		 Three consecutive frames from a testing sequence in the DISK dataset. First row: the results of ACAP. As shown in the second column, the enlarged wrinkles are different from the previous and the next frames.
		 This causes jumping in the animation.
		 Second row: the consistent results obtained via TS-ACAP feature, demonstrating that our TS-ACAP representation ensures the temporal coherence. 
	}
	\label{fig:jump_acap}
\end{figure}
\begin{table}[tb]
	\renewcommand\arraystretch{1.5}
	\fontsize{7.5}{9}\selectfont
	\caption{Comparison of RMSE between synthesized shapes and ground truth with different networks, \ie without temporal modules, with RNN, with LSTM and ours with the Transformer network.}
	\label{table:transformer_compare}
	\centering
	\begin{tabular}{cccccc}
		\toprule[1.2pt]
		Dataset        & TSHIRT      & PANTS    &  	SKIRT   & SHEET       & DISK \\ \hline
		WO Transformer &  0.00909     & 0.01142   &    0.00831 &   0.00739    &  0.0427    \\ \hline
		With RNN  &  0.0435      &  0.0357   & 0.0558     &   0.0273     & 0.157      \\ \hline
		With LSTM  &  0.0351      &  0.0218   &  0.0451    &  0.0114      &  0.102     \\ \hline
		With Transformer         &  \textbf{0.00546}     & \textbf{0.00663}  &    \textbf{0.00685} &   \textbf{0.00585} & \textbf{0.0216} \\ 
		\bottomrule[1.2pt]
	\end{tabular}
\end{table} 
\begin{figure}[tb]
 	\centering
    \setlength{\tabcolsep}{0.0cm}  
    \renewcommand\arraystretch{-1.9}
 	\begin{tabular}{>{\centering\arraybackslash}m{0.08\linewidth}>{\centering\arraybackslash}m{0.18\linewidth}>{\centering\arraybackslash}m{0.18\linewidth}>{\centering\arraybackslash}m{0.18\linewidth}>{\centering\arraybackslash}m{0.18\linewidth}>{\centering\arraybackslash}m{0.18\linewidth}}
 		\rotatebox{90}{\small (a) Input}& 
		\includegraphics[width=\linewidth, trim=5 5 5 5,clip]{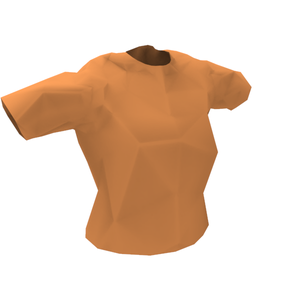} &
		\includegraphics[width=\linewidth, trim=5 5 5 5,clip]{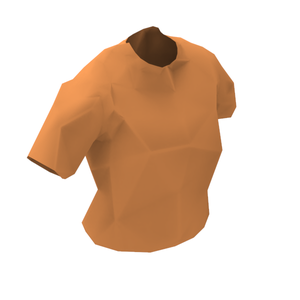} &
		\includegraphics[width=\linewidth, trim=5 5 5 5,clip]{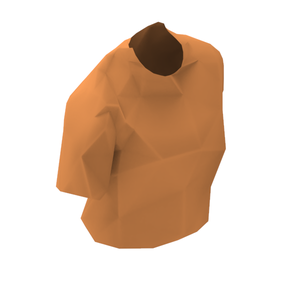} &
		\includegraphics[width=\linewidth, trim=5 5 5 5,clip]{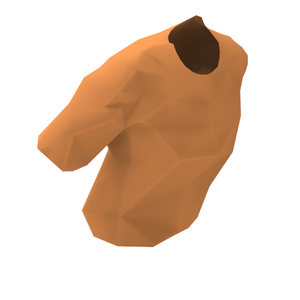} &
		\includegraphics[width=\linewidth, trim=5 5 5 5,clip]{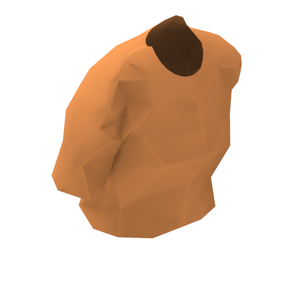}   
 		\\
 		 \rotatebox{90}{\small (b) EncDec} &
 		\includegraphics[width=\linewidth, trim=5 5 5 5,clip]{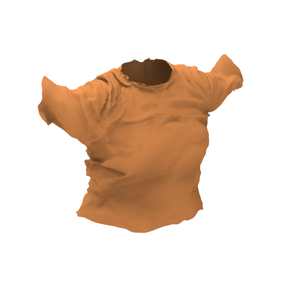} &
 		\includegraphics[width=\linewidth, trim=5 5 5 5,clip]{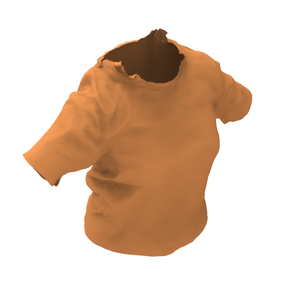} &
 		\includegraphics[width=\linewidth, trim=5 5 5 5,clip]{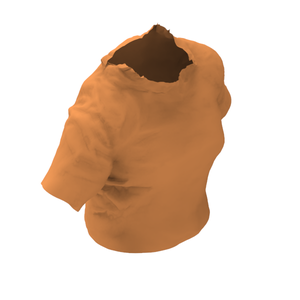} &
 		\includegraphics[width=\linewidth, trim=5 5 5 5,clip]{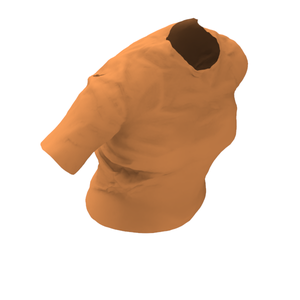} &
 		\includegraphics[width=\linewidth, trim=5 5 5 5,clip]{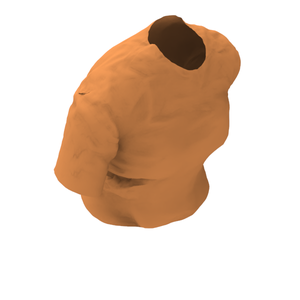} 
 		\\
 		 \rotatebox{90}{\small (c) RNN} &
 		\includegraphics[width=\linewidth, trim=5 5 5 5,clip]{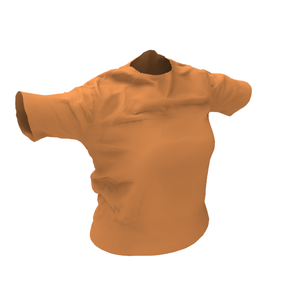} &
 		\includegraphics[width=\linewidth, trim=5 5 5 5,clip]{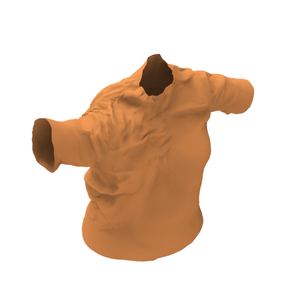} &
 		\includegraphics[width=\linewidth, trim=5 5 5 5,clip]{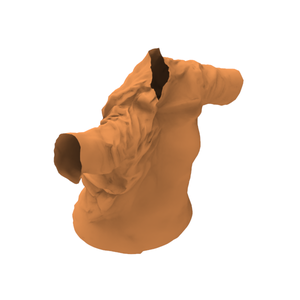} &
 		\includegraphics[width=\linewidth, trim=5 5 5 5,clip]{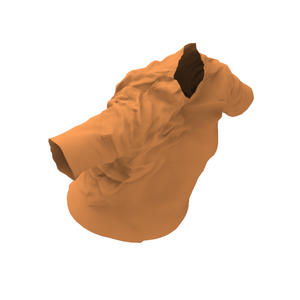} &
 		\includegraphics[width=\linewidth, trim=5 5 5 5,clip]{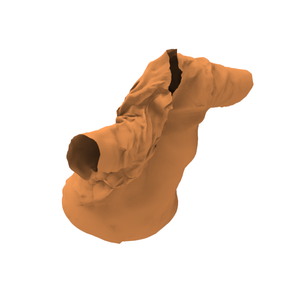} 
	 	\\
	 	\rotatebox{90}{\small (d) LSTM}&
	 	\includegraphics[width=\linewidth, trim=5 5 5 5,clip]{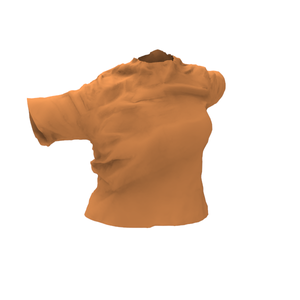}&
	 	\includegraphics[width=\linewidth, trim=5 5 5 5,clip]{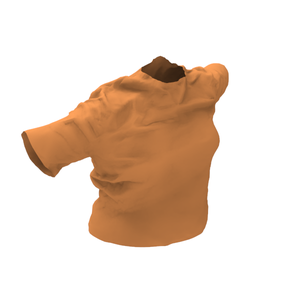}&
	 	\includegraphics[width=\linewidth, trim=5 5 5 5,clip]{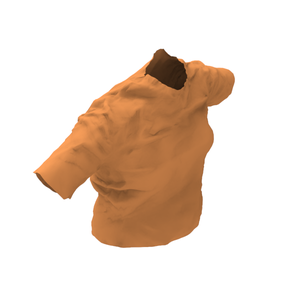}&
	 	\includegraphics[width=\linewidth, trim=5 5 5 5,clip]{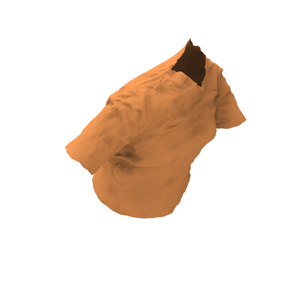}&
	 	\includegraphics[width=\linewidth, trim=5 5 5 5,clip]{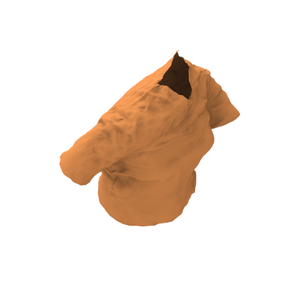} 
 		\\
 		 \rotatebox{90}{\small (e) Ours}& 
 		\includegraphics[width=\linewidth, trim=5 5 5 5,clip]{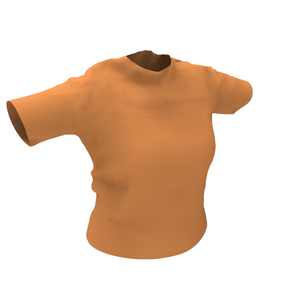}&
 		\includegraphics[width=\linewidth, trim=5 5 5 5,clip]{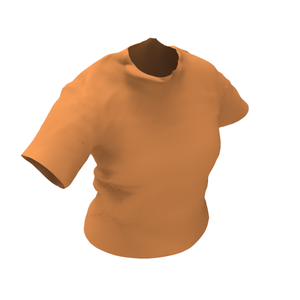}&
 		\includegraphics[width=\linewidth, trim=5 5 5 5,clip]{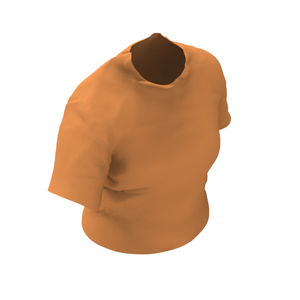}&
 		\includegraphics[width=\linewidth, trim=5 5 5 5,clip]{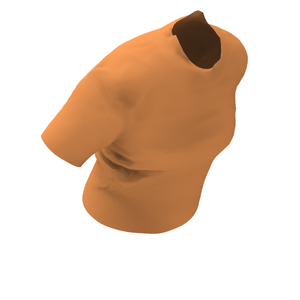}&
 		\includegraphics[width=\linewidth, trim=5 5 5 5,clip]{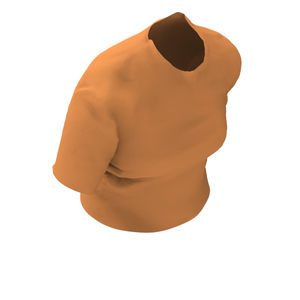} 
 		\\ 
 		  \rotatebox{90}{\small (f) GT}&
 		\includegraphics[width=\linewidth, trim=5 5 5 5,clip]{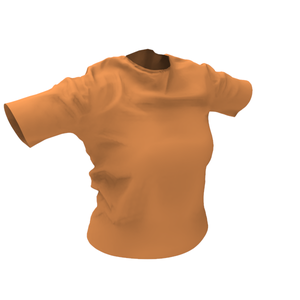}&
 		\includegraphics[width=\linewidth, trim=5 5 5 5,clip]{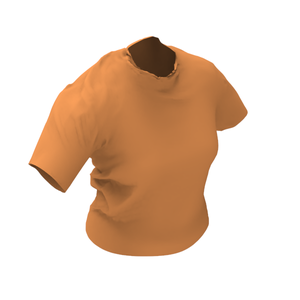}&
 		\includegraphics[width=\linewidth, trim=5 5 5 5,clip]{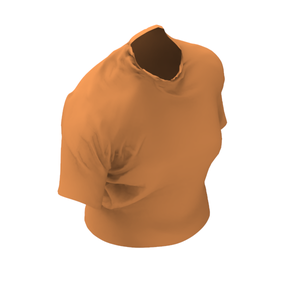}&
 		\includegraphics[width=\linewidth, trim=5 5 5 5,clip]{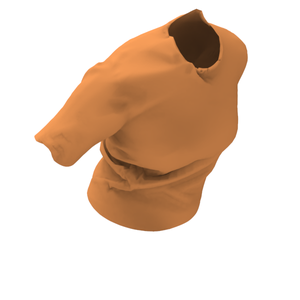}&
 		\includegraphics[width=\linewidth, trim=5 5 5 5,clip]{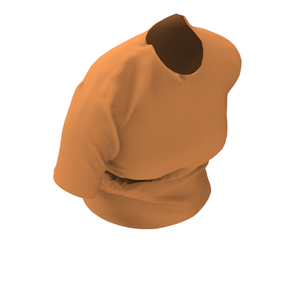} 
 	\end{tabular} 
 	\caption{The evaluation of the Transformer network in our model for wrinkle synthesis.
 		From top to bottom we show (a) %
 		\gl{input coarse mesh with physical simulation}
 		(b) the results with an encoder-decoder \YL{dropping out temporal modules}, (c) the results with RNN \cite{chung2014empirical}, (d) the results with LSTM \cite{hochreiter1997long}, (e) ours, and (f) the ground truth generated by PBS.}
 	\label{fig:transformer_w_o_tshirt}
 \end{figure}   

\subsection{\YL{Evaluation of} Network Components}
We evaluate the effectiveness of our network components for two aspects: the \YL{capability} of the TS-ACAP feature and the \YL{capability} of the Transformer network. 
We evaluate our method qualitatively and quantitatively on different datasets.

\textbf{Feature Representation Evaluation}.
To verify the effectiveness of our TS-ACAP feature, we compare per-vertex position errors to other features to evaluate the generated shapes in different datasets quantitatively. 
We compare our method using TS-ACAP feature with our transduction methods using 3D vertex coordinates and ACAP, with network layers and parameters adjusted accordingly to optimize performance alternatively.
The details of numerical comparison are shown in Table \ref{table:feature_compare}.
ACAP and TS-ACAP show quantitative improvements than 3D coordinates.  
In Fig. \ref{fig:ablationstudy_coordiniates_skirt}, we exhibit several compared examples of animated skirts of coordinates and TS-ACAP. 
\YL{The results using coordinates show rough appearance, unnatural deformation and some artifacts, 
 I can't really see the two circles?
especially in the highlighted regions with details shown in the close-ups.} Our results with TS-ACAP are more similar to the ground truth than the ones with coordinates. 
ACAP has the problem of temporal inconsistency, thus the results are shaking or jumping frequently. 
\YL{Although the use of the Transformer network can somewhat mitigate this issue, such artifacts can appear even with the Transformer.}
\YL{Fig.~\ref{fig:jump_acap} shows} three consecutive frames from a testing sequence in the DISK dataset.
Results with TS-ACAP show more consistent wrinkles than the ones with ACAP thanks to the temporal constraints.

\textbf{Transformer Network Evaluation}.
We also evaluate the impact of the Transformer network in our pipeline. 
We compare our method to an encoder-decoder network dropping out the temporal modules, our pipeline with the recurrent neural network (RNN) and with the long short-term memory (LSTM) \YL{module}.
An example of T-shirts is given in Fig. \ref{fig:transformer_w_o_tshirt}, \YL{showing} 5 frames in order.
The results without any temporal modules show artifacts on the sleeves and neckline since these places have strenuous \YL{forces}. %
The models using RNN and LSTM stabilize the sequence via eliminating dynamic and detailed deformation, but all the results keep wrinkles on the chest from the initial state\YL{, lacking rich dynamics.}
Besides, they are not able to generate stable and realistic garment animations \YL{that look similar to} the ground truth,
\YL{while} \YL{our} method with the Transformer network \YL{apparently} improves the temporary stability, \YL{producing results close to the ground truth.}
We also quantitatively evaluate the performance of the Transformer network \YL{in our method} via per-vertex error. 
As shown in Table \ref{table:transformer_compare}, the RMSE of our model \YL{is} smaller than the other models.

\section{Conclusion and Future Work}\label{sec:conclusion}
In this paper, we introduce a novel algorithm for synthesizing robust and realistic cloth animations via deep learning.
To achieve this, we propose a geometric deformation representation named TS-ACAP which well embeds the details and ensures the temporal consistency.
\YL{Benefiting} from \YL{the} deformation-based feature, there is no explicit requirement of tracking between coarse and fine meshes in our algorithm. 
We also use the Transformer network based on attention mechanisms to map the coarse TS-ACAP to fine TS-ACAP, maintaining the stability of our generation.
Quantitative and qualitative results reveal that our method can synthesize realistic-looking wrinkles in various datasets, such as draping tablecloth, tight or \YL{loose} garments dressed on human bodies, etc. 
 
Since our algorithm synthesizes \YL{details} based on the coarse meshes, the time for coarse simulation is unavoidable.
Especially for tight garments like T-shirts and pants, the collision solving phase is time-consuming.
In the future, we intend to generate coarse sequences for tight cloth via skinning-based methods in order to reduce the computation for our pipeline.
Another limitation is that our current network is not able to deal with all kinds of garments with different topology.
\newpage
\bibliographystyle{IEEEtran}
\bibliography{reference}